\documentclass[runningheads]{llncs}

\usepackage[year=2026]{eccv}

\usepackage{eccvabbrv}

\usepackage{graphicx}
\usepackage{booktabs}

\usepackage[accsupp]{axessibility}  

\usepackage{hyperref}

\usepackage{orcidlink}

\usepackage{import}
\usepackage{siunitx}
\usepackage{acronym} 
\usepackage{multirow}
\usepackage{xfp} 
\usepackage{pifont} 

\newcommand{\MDGEOsmallP}{0.642}
\newcommand{\MDGEOsmallR}{0.638}
\newcommand{\MDGEOsmallF}{0.640}
\newcommand{\MDTOPOsmallP}{0.570}
\newcommand{\MDTOPOsmallR}{0.553}
\newcommand{\MDTOPOsmallF}{0.562}

\newcommand{\MDGEObigP}{0.772}
\newcommand{\MDGEObigR}{0.765}
\newcommand{\MDGEObigF}{0.768}
\newcommand{\MDTOPObigP}{0.714}
\newcommand{\MDTOPObigR}{0.691}
\newcommand{\MDTOPObigF}{0.702}

\newcommand{\MDIoUsmall}{0.482} 
\newcommand{\MDIoUbig}{0.656} 

\newcommand{\BGFGEOsmallP}{0.447}
\newcommand{\BGFGEOsmallR}{0.469}
\newcommand{\BGFGEOsmallF}{0.458}
\newcommand{\BGFTOPOsmallP}{0.353}
\newcommand{\BGFTOPOsmallR}{0.337}
\newcommand{\BGFTOPOsmallF}{0.345}
\newcommand{\BGFGEObigP}{0.585}
\newcommand{\BGFGEObigR}{0.607}
\newcommand{\BGFGEObigF}{0.596}
\newcommand{\BGFTOPObigP}{0.504}
\newcommand{\BGFTOPObigR}{0.488}
\newcommand{\BGFTOPObigF}{0.496}
\newcommand{\BGFIoUsmall}{0.354} 
\newcommand{\BGFIoUbig}{0.518} 

\newcommand{\LGNNGEOsmallPglobal}{0.701}
\newcommand{\LGNNGEOsmallRglobal}{0.451}
\newcommand{\LGNNGEOsmallFglobal}{0.545}
\newcommand{\LGNNTOPOsmallPglobal}{0.307}
\newcommand{\LGNNTOPOsmallRglobal}{0.281}
\newcommand{\LGNNTOPOsmallFglobal}{0.327}
\newcommand{\LGNNIoUsmallglobal }{0.384}
\newcommand{\LGNNGEObigPglobal}{0.827}
\newcommand{\LGNNGEObigRglobal}{0.531}
\newcommand{\LGNNGEObigFglobal}{0.642}
\newcommand{\LGNNTOPObigPglobal}{0.395}
\newcommand{\LGNNTOPObigRglobal}{0.363}
\newcommand{\LGNNTOPObigFglobal}{0.378}
\newcommand{\LGNNIoUbigglobal }{0.529}

\newcommand{\BGFGEOsmallPglobal}{0.419}
\newcommand{\BGFGEOsmallRglobal}{0.312}
\newcommand{\BGFGEOsmallFglobal}{0.352}
\newcommand{\BGFTOPOsmallPglobal}{0.236}
\newcommand{\BGFTOPOsmallRglobal}{0.158}
\newcommand{\BGFTOPOsmallFglobal}{0.189}
\newcommand{\BGFIoUsmallglobal}{0.280}
\newcommand{\BGFGEObigPglobal}{0.548}
\newcommand{\BGFGEObigRglobal}{0.406}
\newcommand{\BGFGEObigFglobal}{0.459}
\newcommand{\BGFTOPObigPglobal}{0.341}
\newcommand{\BGFTOPObigRglobal}{0.233}
\newcommand{\BGFTOPObigFglobal}{0.276}
\newcommand{\BGFIoUbigglobal}{0.409}

\newcommand{\MDGEOsmallPglobal}{0.575}
\newcommand{\MDGEOsmallRglobal}{0.598}
\newcommand{\MDGEOsmallFglobal}{0.584}
\newcommand{\MDTOPOsmallPglobal}{0.425}
\newcommand{\MDTOPOsmallRglobal}{0.383}
\newcommand{\MDTOPOsmallFglobal}{0.401}
\newcommand{\MDIoUsmallglobal}{0.454}
\newcommand{\MDGEObigPglobal}{0.696}
\newcommand{\MDGEObigRglobal}{0.752}
\newcommand{\MDGEObigFglobal}{0.720}
\newcommand{\MDTOPObigPglobal}{0.585}
\newcommand{\MDTOPObigRglobal}{0.536}
\newcommand{\MDTOPObigFglobal}{0.557}
\newcommand{\MDIoUbigglobal}{0.668}

\newcommand{\MDTOPOsmallPLCnoGL}{0.544}
\newcommand{\MDTOPOsmallRLCnoGL}{0.508}
\newcommand{\MDTOPOsmallFLCnoGL}{0.526}

\newcommand{\MDGEOsmallPLCnoGL}{0.609}
\newcommand{\MDGEOsmallRLCnoGL}{0.634}
\newcommand{\MDGEOsmallFLCnoGL}{0.620}

\newcommand{\MDIoUsmallLCnoGL}{0.478}

\newcommand{\MDTOPOsmallPnoLCnoGL}{0.478}
\newcommand{\MDTOPOsmallRnoLCnoGL}{0.471}
\newcommand{\MDTOPOsmallFnoLCnoGL}{0.484}

\newcommand{\MDGEOsmallPnoLCnoGL}{0.413}
\newcommand{\MDGEOsmallRnoLCnoGL}{0.625}
\newcommand{\MDGEOsmallFnoLCnoGL}{0.498}

\newcommand{\MDIoUsmallnoLCnoGL}{0.286}

\newcommand{\MDTOPOsmallPnoVAE}{0.428}
\newcommand{\MDTOPOsmallRnoVAE}{0.408}
\newcommand{\MDTOPOsmallFnoVAE}{0.418}

\newcommand{\MDGEOsmallPnoVAE}{0.481}
\newcommand{\MDGEOsmallRnoVAE}{0.562}
\newcommand{\MDGEOsmallFnoVAE}{0.519}

\newcommand{\MDIoUsmallnoVAE}{0.417}

\acrodef{VAE}{Variational Autoencoder}
\acrodef{LDM}{Latent Diffusion Model}

\begin{document}

\title{MapDreamer: Aerial Imagery Conditioned Latent Diffusion for Lane-Level Map Generation}
\titlerunning{MapDreamer: LDM Map Generation from Aerial Imagery}

\author{Julian Brandes\inst{1,2}\orcidlink{0009-0008-7732-6937} \and
Philipp Crocoll\inst{2}\orcidlink{0000-0002-4558-0307} \and
Wolfram Burgard\inst{1}\orcidlink{0000-0002-5680-6500}}

\authorrunning{J.~Brandes et al.}

\institute{
Department CSAI at University of Technology Nuremberg, Nuremberg, Germany\\
\and
Robert Bosch GmbH, Stuttgart\\
}
\maketitle
\begin{abstract}
High definition map generation is essential for autonomous driving, yet remains a labor-intensive process at scale.
We present MapDreamer, a generative diffusion model that synthesizes lane-level vector maps with explicit topology directly from a single aerial image.
MapDreamer learns a compact latent representation of lane centerlines and their topological relations using a variational autoencoder and predicts graphs with a transformer-based latent diffusion model.
To align generated maps with the observed scene, we condition each denoising step on dense aerial features injected through cross-attention.
To handle the varying number of lanes across scenes, we propose a lane cardinality module paired with background ghost lane latents, a learned buffer that prevents slot collapse during diffusion.
Furthermore, we introduce a sliding-window global graph aggregation strategy that stitches local tiles into city-scale maps while preserving connectivity through encoded lane boundaries. Experiments on Urban\-Lane\-Graph derived from Argoverse 2 show improved geometric and topological fidelity over non-generative baselines.

\end{abstract}    
\section{Introduction}
\label{sec:intro}

\begin{figure}
    \centering
\begingroup%
  \makeatletter%
  \providecommand\color[2][]{%
    \errmessage{(Inkscape) Color is used for the text in Inkscape, but the package 'color.sty' is not loaded}%
    \renewcommand\color[2][]{}%
  }%
  \providecommand\transparent[1]{%
    \errmessage{(Inkscape) Transparency is used (non-zero) for the text in Inkscape, but the package 'transparent.sty' is not loaded}%
    \renewcommand\transparent[1]{}%
  }%
  \providecommand\rotatebox[2]{#2}%
  \newcommand*\fsize{\dimexpr\f@size pt\relax}%
  \newcommand*\lineheight[1]{\fontsize{\fsize}{#1\fsize}\selectfont}%
  \ifx\svgwidth\undefined%
    \setlength{\unitlength}{345.82677165bp}%
    \ifx\svgscale\undefined%
      \relax%
    \else%
      \setlength{\unitlength}{\unitlength * \real{\svgscale}}%
    \fi%
  \else%
    \setlength{\unitlength}{\svgwidth}%
  \fi%
  \global\let\svgwidth\undefined%
  \global\let\svgscale\undefined%
  \makeatother%
  \begin{picture}(1,0.3852459)%
    \lineheight{1}%
    \setlength\tabcolsep{0pt}%
    \put(0,0){\includegraphics[width=\unitlength,page=1]{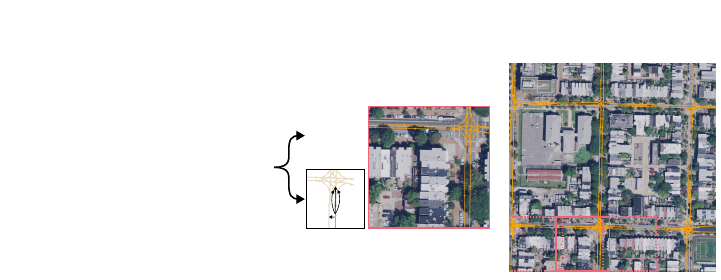}}%
    \put(0.46477898,0.04125211){\makebox(0,0)[t]{\lineheight{1.20000005}\smash{\begin{tabular}[t]{c}Graph\\Topology\end{tabular}}}}%
    \put(0,0){\includegraphics[width=\unitlength,page=2]{abstract_figure.pdf}}%
    \put(0.46532826,0.28859427){\makebox(0,0)[t]{\lineheight{1.20000005}\smash{\begin{tabular}[t]{c}Vectorized Lane\\Geometry\end{tabular}}}}%
    \put(0.85121669,0.36549616){\makebox(0,0)[t]{\lineheight{1.20000005}\smash{\begin{tabular}[t]{c}\textit{Global Graph}\\\textit{Aggregation}\end{tabular}}}}%
    \put(0.35131403,0.34813201){\makebox(0,0)[t]{\lineheight{1.20000005}\smash{\begin{tabular}[t]{c}\textit{Local Graph Prediction}\end{tabular}}}}%
    \put(0,0){\includegraphics[width=\unitlength,page=3]{abstract_figure.pdf}}%
    \put(0.29917316,0.14785618){\color[rgb]{1,1,1}\makebox(0,0)[t]{\smash{\begin{tabular}[t]{c}MapDreamer\end{tabular}}}}%
    \put(0,0){\includegraphics[width=\unitlength,page=4]{abstract_figure.pdf}}%
  \end{picture}%
\endgroup%

    \caption{We propose MapDreamer for lane-level graph generation from aerial imagery. For each input tile, vectorized lane geometry is predicted along with the corresponding graph topology. For city-scale map prediction, we aggregate a global graph through overlapping imagery tiles. The figure displays inference outputs of MapDreamer.}
    \label{fig:overview_task}
\end{figure}

High definition, lane-level maps encode road geometry with high accuracy along with directed lane connectivity and thus support planning, simulation, and large-scale evaluation of autonomous driving tasks. 
Producing such maps at scale remains a challenge, one of the main factors being the costly and slow manual annotation. The use of automated, offline pipelines can enable low-cost maps that still support high-level autonomous driving.
Aerial imagery provides a complementary sensing modality for lane-level map construction through broad coverage and visual cues such as lane markings, curb geometry, and intersection layout.
However, converting aerial pixels into reliable and accurate lane-level graphs is challenging. 
While many methods exist aiming to generate global graphs on road-level from aerial image inputs~\cite{mattyus_deeproadmapper_2017,bastani_roadtracer_2018,vedaldi_sat2graph_2020,tan_vecroad_2020,xu_rngdet_2023,yin_towards_2025} enabling navigation tasks, relatively few extract lane-level information required for high-level functions~\cite{he_lane-level_2022,buchner_learning_2023,blayney_bezier_2024}.

In the past, methods based on pixel-wise segmentation leading to rasterized outputs built the foundation of graph extraction from aerial imagery. However, rasterized outputs require brittle post-processing to recover topology, where small errors can disconnect long-range routes motivating the recent shift to methods based on vectorized geometries~\cite{bastani_roadtracer_2018, li_topological_2019, liu_vectormapnet_2023}.
While creating local graphs for a scene is important for precise mapping of complex intersections, creating consistent global graphs is equally important. Overcoming this local-to-global gap is challenging since lane graphs are discrete and highly structured, yet the evidence in imagery is local and incomplete. 

In this paper, we propose MapDreamer, a generative latent diffusion model for local and global lane graph generation conditioned on aerial imagery. 
We represent a lane map as a set of lane centerlines with explicit relations and train a variational autoencoder that maps this structured representation into a compact latent space.
This latent formulation compresses variable sized lane graphs into a fixed-dimensional representation, which reduces diffusion model complexity. At the same time, the decoder imposes geometric and topological plausibility, so sampling in latent space is implicitly regularized toward valid lane topology observed in real scenes.
We train a diffusion model in this latent space conditioned on aerial features extracted from a vision backbone.
We use a learned lane cardinality module to handle the large variation in lane density across scenes, from sparse rural roads to dense urban intersections. To avoid recall degradation, we further introduce \textit{ghost lane latents} as background slots that act as capacity buffers, allowing the model to generate additional lanes when supported by evidence while reducing duplicated or redundant segments. 
For city-scale graph generation, we formulate a sliding-window inference strategy where tiles are conditioned on already predicted lane centerlines from neighboring tiles in order to respect boundary context.
\Cref{fig:overview_task} summarizes the local graph prediction and the tile-stitching strategy for city-scale reconstruction.
\subsubsection{Contributions.}
We propose MapDreamer, a latent diffusion framework for lane-level map generation from aerial imagery that captures geometric structure and directed connectivity in a learned latent space. 
We introduce \textit{ghost latents} as a background mechanism along with a lane cardinality module, enabling flexible generation under uncertain lane counts, improving stability across scenes with varying lane cardinality. 
We develop a sliding window inference strategy that maintains topological continuity across tile boundaries for scalable city-wide deployment. Finally, we provide comprehensive evaluation demonstrating improved topological consistency beyond standard polyline detection metrics, validated through connectivity-aware measures on large-scale lane graphs.
\section{Related Work}
\label{sec:related_work}

\subsection{Road Graph Extraction from Aerial Imagery}
Early road-level learning-based pipelines commonly predict segmentation maps and convert them into networks via heuristic graph recovery, which is sensitive to gaps caused by low-contrast markings or occlusions~\cite{cheng_automatic_2017}. 
DeepRoadMapper~\cite{mattyus_deeproadmapper_2017} traces road geometry to recover connectivity, RoadTracer formulates map construction as iterative graph expansion guided by a learned decision function~\cite{bastani_roadtracer_2018}. Sat2Graph encodes graphs into a tensor representation, enabling dense prediction that still supervises graph structure~\cite{vedaldi_sat2graph_2020}. More recent methods aim to predict complete graphs with fewer handcrafted steps, including transformer-based formulations such as RNGDet~\cite{xu_rngdet_2022} and RNGDet++~\cite{xu_rngdet_2023} for large-scale road network detection from aerial imagery, SAM-Road++~\cite{yin_towards_2025} adapts the Segment Anything Model as a backbone for road graph extraction.

\subsection{Lane-Level Graph Generation from Aerial Imagery}
Lane-level mapping requires finer geometry and directed relations, such as successor and lateral connectivity, which increases ambiguity at complex junctions.
LaneExtraction~\cite{he_lane-level_2022} extracts lane-level maps from aerial imagery using a staged pipeline that predicts non-intersection lane segmentation and per-pixel directions, and then converts these dense outputs into a directed graph with dedicated handling for turning lanes at intersections.
The UrbanLaneGraph dataset~\cite{buchner_learning_2023} emphasizes large-scale lane graph estimation from aerial tiles and introduces connectivity-focused evaluation.
LaneGNN~\cite{buchner_learning_2023} predicts successor lane graphs from oriented aerial crops by placing a virtual agent at the bottom center of the crop.
From a predicted lane centerline likelihood map and an ego-lane segmentation mask of reachable lanes, it samples proposal nodes, connects nearby pairs to form a candidate directed graph, and scores edges with a direction-aware GNN~\cite{buchner_learning_2023}.
BGFormer proposes a Bézier graph representation and a transformer-based lane graph generator from aerial images, highlighting the benefit of structured curve parameterizations for smoother, drivable lane networks~\cite{blayney_bezier_2024}.

\subsection{Online Map Generation Using Map Priors}
A related line of work constructs local vectorized maps online from vehicle sensors. Many approaches learn BEV features from multi-view images and decode sets of vector elements~\cite{li_hdmapnet_2022,liu_vectormapnet_2023,liao_maptrv2_2024}, temporal models further improve stability and range~\cite{yuan_streammapnet_2024}. Mask2Map injects mask based reasoning into vector decoding to better capture diverse shapes~\cite{leonardis_mask2map_2025}. These methods target ego-centric perception under partial observability and typically output collections of map elements, while explicit directed lane-graph connectivity is often neglected, differing from our overhead setting.
Several works also incorporate aerial imagery as map priors. SatForHDMap studies how satellite map tiles can complement onboard sensors through alignment and cross-attention style fusion~\cite{gao_complementing_2024}, SMART scales map prior learning using standard definition maps and satellite inputs to improve topology reasoning~\cite{ye_smart_2025}. These results support the view that overhead imagery can provide strong priors for lane topology, even when the final task is sensor based.

\subsection{Diffusion Models for Maps}
Diffusion models have become the state-of-the-art tool for conditional generation, with denoising performed either in pixel space~\cite{ho_denoising_2020} or in a learned autoencoder latent space for improved efficiency and performance~\cite{rombach_high-resolution_2022}. Further, transformer backbones have been adopted for diffusion, demonstrating strong scaling behavior~\cite{peebles_scalable_2023}.

Recent work has adapted diffusion models to vectorized map perception in online map generation.
MapDiffusion~\cite{monninger_mapdiffusion_2025} replaces deterministic query refinement with iterative denoising of randomly initialized polyline queries conditioned on a BEV latent grid, predicting an ego-centric local map as sets of polylines for lane dividers, lane boundaries, and pedestrian crossings. 
The work emphasizes uncertainty in the perceived map, quantified via sample variance, and refines predictions through rasterized sample aggregation~\cite{monninger_mapdiffusion_2025}.
Our setting differs in that we infer directed lane graphs with explicit connectivity from overhead imagery and target large-area reconstruction by stitching overlapping tiles. This motivates our diffusion model in a learned lane-graph latent space that regularizes geometry and adjacency jointly.

LaneDiffusion~\cite{wang_zijie_lanediffusion_2025} applies diffusion at the BEV feature level by constructing diffusion targets through lane-prior injection, vectorized centerlines and topology are predicted using a downstream graph decoder.
For overhead imagery, Gu \etal generate rasterized road images with conditional diffusion under geospatial context and then convert them into networks via post-processing~\cite{gu_generating_2024}. We avoid raster intermediates and directly generate vector lane graphs with directed connectivity.

\subsubsection{Vectorized Generative Simulation.}
Scenario generation provides another perspective on structured map generation. Scenario Dreamer proposes vectorized latent diffusion for generating lane graphs together with other scene elements for driving simulation, showing that latent diffusion over structured vector primitives can produce high quality simulation scenarios from a real-world scene distribution~\cite{rowe_scenario_2025}. 
While its goal is synthetic environment generation rather than aerial map extraction, it supports the design choice of modeling structured lane graphs with latent diffusion.

\section{MapDreamer}
\label{sec:mapdreamer}
We propose MapDreamer, a latent space generative method for lane-level map prediction from aerial imagery, inspired by the driving scenario simulator Scenario Dreamer~\cite{rowe_scenario_2025}. 

\subsection{Scenario Dreamer}
The work by Rowe \etal aims to generate simulation scenarios consisting of a lane graph and agent bounding boxes, and models agent behavior in closed loop. For initial scene generation, it follows a two-stage design. First, a \ac{VAE} is used to learn a latent representation of vectorized centerlines and their topology along with traffic participant bounding boxes. The encoded agent and lane latents are used in the second stage to train a transformer-based latent diffusion model that is able to generate unseen scenarios sampled from the learned distributions of the training scenes.

\subsection{MapDreamer Overview}
\begin{figure}[t]
    \centering
    \def\svgwidth{\linewidth} 
    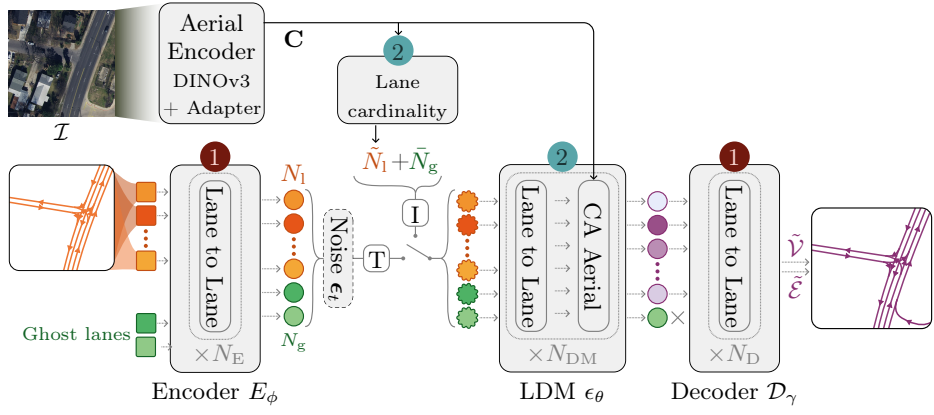
    \caption{Overview of MapDreamer training and inference at diffusion step $t$. 
    Stage one trains the VAE encoder $E_\phi$ and decoder $\mathcal{D}_\gamma$, stage two freezes $E_\phi$ and trains the LDM noise predictor $\epsilon_\theta$ and lane cardinality module $\pi_\psi$.
    Training (T) uses the ground truth number of latents $N_\mathrm{l}$ plus a random number of $N_\mathrm{g}$ ghost latents, while inference (I) uses the predicted number $\tilde N_\mathrm{l}$ of latents plus a fixed number $\bar N_\mathrm{g}=5$ of ghost latents.} 
    \label{fig:mapdreamer_architecture}
\end{figure}

We adapt the two-stage diffusion training and reformulate it for fully conditional lane graph reconstruction from aerial imagery. Instead of generating unseen simulation scenarios, we learn a conditional generative model to predict maps from the learned distribution, enabling geometrically accurate and topologically consistent maps.
\Cref{fig:mapdreamer_architecture} illustrates the training and inference strategy of the MapDreamer architecture.

\subsection{Scene Representation}
Given an aerial tile \(\mathcal{I}\in\mathbb{R}^{H\times W\times C}\) covering the area \(\Omega\subset\mathbb{R}^2\) of size \(76.8\,\mathrm{m}\times 76.8\,\mathrm{m}\) at \(512\times512\) pixels, we predict a lane graph \(\tilde{\mathcal{G}}=\{\tilde{\mathcal{V}},\tilde{\mathcal{E}}\}\) corresponding to the ground truth graph \(\mathcal{G}=\{\mathcal{V},\mathcal{E}\}\).
The node set \(\mathcal{V}=\{v_i\}_{i=1}^{N_\mathrm{l}}\) represents vectorized lane segments, where each lane is parameterized by its centerline polyline \(v_i=[\mathbf p_{i,0},\dots,\mathbf p_{i,n-1}]\) with \(n=20\) two-dimensional points \(\mathbf p_{i,j}\in\Omega\).
Edges \(\mathcal{E}\) encode lane connectivity and are represented by four adjacency matrices \(\mathbf{A}\in\{0,1\}^{N_\mathrm{l}\times N_\mathrm{l}\times 4}\) for successor, predecessor, left, and right relations~\cite{rowe_scenario_2025}.
We merge degree-two edges during preprocessing, so lanes divide only at merges, splits, or at terminations without successor or predecessor in \(\mathcal{G}\).

\subsection{Model Architecture}
\label{subsec:mapdreamer_model_architecture}
To align the generative denoising process with the real-world scenario, we condition the diffusion model on aerial image features.
We extract dense visual tokens from pretrained DINOv3~\cite{simeoni_dinov3_2025} vision foundation model features, reducing the task-specific feature learning required for aerial tiles. 
The diffusion transformer attends to the aerial tokens through dense cross-attention at every denoising step, allowing the predicted lane latents to align with the image evidence.

Our goal is to sample lane graphs from the conditional distribution \(p(\mathcal{G}\mid\mathcal{I})\), which represents the probability of a lane graph configuration $\mathcal{G}$ given an aerial image tile $\mathcal{I}$. To achieve this, we learn an approximation $\tilde p(\mathcal{G} \mid \mathcal{I})\approx p(\mathcal{G} \mid \mathcal{I})$ using a diffusion model, benefiting from its capability to capture complex, multimodal spatial distributions.

Unlike the scenario generation in Scenario Dreamer~\cite{rowe_scenario_2025}, lane cardinality is unknown at inference for map prediction.
Across our dataset, the number of lane segments varies strongly across tiles, ranging from simple road segments with only one or two lanes to dense intersections containing dozens of lane segments.
Empirically, we found the use of a lane existence head and a fixed-size set of lane queries to be brittle for lane graph prediction because setting the query budget high enough for complex scenes leaves many surplus slots in simple scenes, see \cref{subsec:ablations}. Often, multiple lane queries collapse into near-identical hypotheses and produce stacked, overlapping lane segments.
We therefore introduce a \textit{lane cardinality module} to estimate the number of lanes $\tilde N_\mathrm{l}$ in the image tile $\mathcal{I}$ used to initialize the primary latent lane queries during inference.

Since $\tilde {N_\mathrm{l}}$ is an estimate, strictly relying on it creates a bottleneck for recall.
To address this, we propose \textit{Ghost Lane Latents} as a fixed set of background lane embeddings encoded via the autoencoder at initialization.
During latent diffusion training, we append a random number of ghost latents $N_\mathrm{g}\in\{0,\dots,N_\mathrm{g}^{\max}=8\}$ to the ground truth lane latents, making the diffusion process robust to both under- and over-capacity initializations.
During inference, we add a fixed number of $\bar N_{\mathrm{g}}=5$ queries to the estimated lane count $\tilde {N_\mathrm{l}}$, chosen based on the empirical lane-cardinality estimation error (MAE: 3.34).
These ghost latents serve as a reasoning buffer and allow the diffusion model to dynamically adjust the required number of lanes for the tile by suppressing excess queries or activating ghost slots to recover missed lanes from the initial estimation.

\subsection{\acl{VAE}}
\label{subsec:vae}
In the first stage, we train a \ac{VAE} that maps structured vector lane graphs to a compact, continuous latent space. This latent space enables efficient diffusion-based modeling, while the fixed decoder provides a structural prior when mapping latents back to lane geometry and their connectivity. We adopt the autoencoder design of Scenario Dreamer~\cite{rowe_scenario_2025}, using a transformer-based encoder $E_\phi$ and decoder $\mathcal{D}_\gamma$.

\subsubsection{Encoder.}
The encoder embeds each of the $N_\mathrm{l}$ lane polylines with a per-lane MLP. Pairwise lane relations derived from the graph edges $\mathcal{E}$ are embedded with an additional connectivity MLP and used as edge features. We use $N_{E_\phi}=2$ lane-to-lane attention blocks~\cite{rowe_scenario_2025} and project the lane embeddings to Gaussian latents of dimension $K_\mathrm{l}$.
\subsubsection{Decoder.}
We sample latents with the reparameterization trick and decode with $\mathcal{D}_\gamma$ to reconstruct lane geometry and connectivity.
Latents are projected to the model dimension and processed with \(N_{\mathcal{D}_\gamma}=2\) lane-to-lane attention blocks. Finally, output heads reconstruct the lane geometry and pairwise connectivity.
The VAE is trained with a \(\beta\)-weighted ELBO
\begin{equation}
\mathcal{L}_{\mathrm{VAE}}
=
\mathcal{L}_{\mathrm{rec}}(\mathcal{G}, \mathcal{D}_\gamma(\mathbf{z}))
+
\beta\,D_{\mathrm{KL}}\!\left(q_\phi(\mathbf{z}\mid \mathcal{G})\,\|\,\mathcal{N}(0,I)\right),
\end{equation}
the approximate posterior $q_\phi(\mathbf{z}\mid \mathcal{G})$ is parameterized as a diagonal Gaussian over per lane latents.
The reconstruction term $\mathcal{L}_{\mathrm{rec}}$ combines an $\ell_1$ loss on lane point coordinates with a cross-entropy loss over discrete pairwise lane connectivity types. We additionally regularize successor continuity by penalizing the $\ell_1$ distance between the predicted endpoint of lane $i$ and the predicted starting point of its successor $j$.
After the VAE training, $E_\phi$ and $\mathcal{D}_\gamma$ are frozen and only the latent diffusion model is optimized in stage two.
\subsection{Vectorized Latent Diffusion Model}
\label{subsec:ldm}
We formulate map generation as a conditional latent diffusion model that denoises Gaussian samples toward lane latents defined by the VAE encoder \(E_\phi\), conditioned on an aerial image tile \(\mathcal{I}\).
Our diffusion target is the conditional distribution over variable cardinality latent lane sets.
Let \(N_\mathrm{l}\) denote the number of lane segments in the tile and \(\mathbf Z=\{\mathbf z_i\}_{i=1}^{N_\mathrm{l}}\) the corresponding set of per-lane latents in the latent space.
We aim to approximate
\begin{equation}
p(\mathbf Z \mid \mathcal I)
=
\sum_{n} p(\mathbf Z \mid N_\mathrm{l}=n,\mathcal I)\,p(N_\mathrm{l}=n\mid \mathcal I)
\end{equation}
through a learned diffusion model \(p_\theta(\mathbf Z \mid \mathcal I)\).
In practice, \(p(N_\mathrm{l}\mid \mathcal I)\) is approximated by the lane cardinality module, and we implement diffusion on a fixed token budget by augmenting the lane count with ghost slots.
Training uses \(N_\mathrm{l}+N_\mathrm{g}\) tokens, while inference uses \(N_{\text{tok}}=\tilde N_\mathrm{l}+\bar N_\mathrm{g}\), where $\bar N_\mathrm{g}=5$ is a small buffer treated as ghost lane tokens.
A lane existence head suppresses unused tokens, yielding a variable-size output set.

We apply a DDPM forward corruption process in latent space and learn the conditional reverse process~\cite{ho_denoising_2020,rombach_high-resolution_2022,rowe_scenario_2025}.
The noise-prediction model \(\epsilon_\theta\) is a DiT-style transformer with timestep conditioning~\cite{peebles_scalable_2023} that operates on lane tokens and predicts the injected noise and is trained with
\begin{equation}
\mathcal{L}_{\mathrm{dm}}
=
\mathbb{E}_{t,\boldsymbol{\epsilon}}
\Big[
\lVert \boldsymbol{\epsilon} - \epsilon_{\theta}(\mathbf{Z}_t, t, \mathbf{c}(\mathcal{I})) \rVert_2^2
\Big],
\end{equation}
where \(\mathbf{c}(\mathcal{I})\) is extracted by the aerial encoder and provided through cross-attention.

\subsubsection{Lane Cardinality Module.}
\label{par:lane_cardinality}
To handle variable lane cardinality, the \textit{lane cardinality module}, implemented as a transformer-based counting head, outputs logits inducing a discrete distribution 
\begin{equation}
    \pi_{\psi}(n \mid \mathcal{I}), \quad n \in \{0,\dots,N^{\max}\}, \quad N^{\max}=100.
\end{equation}
The estimated number of lanes $\tilde {N_\mathrm{l}}$ is the expected value
\begin{equation}
    \tilde N_\mathrm{l} = \Big\lfloor \sum_{n=0}^{N^{\max}} n \,\pi_\psi ( n \mid \mathcal{I}) \Big\rfloor.
\end{equation}
The module is trained using cross-entropy loss on the discrete lane count and an auxiliary expectation regression term with an asymmetric penalty for underestimation.
\subsubsection{Ghost Lane Embedding.}
\label{par:ghost_lanes}
Ghost lane tokens are assigned fixed clean targets $\{\mathbf{g}_j\}_{j=1}^{N_\mathrm{g}}$ obtained by encoding simple out-of-bounds lane polylines through the \ac{VAE} encoder, following the lane permutation ambiguity strategy~\cite{rowe_scenario_2025}.
Denoting the augmented clean set by
\begin{equation}
\tilde{\mathbf{Z}}_0 = [\mathbf{z}_{0,1},\dots,\mathbf{z}_{0,N_\mathrm{l}},\mathbf{g}_1,\dots,\mathbf{g}_{N_\mathrm{g}}],
\end{equation}
we apply the same forward process and noise prediction loss to all tokens.
We associate each token with an existence logit $s_i$, predicted by an additional head trained jointly with the diffusion objective. For a token label $y_i \in \{0,1\}$, where $y_i = 1$ denotes a real lane and $y_i = 0$ denotes a ghost token, we optimize an existence loss
\begin{equation}
\mathcal{L}_{\mathrm{exist}}
=
\mathbb{E}_{t}
\left[
\sum_{i=1}^{N_\mathrm{l}+N_{\mathrm{g}}}
w(t)\,\ell_{\mathrm{bin}}(s_i, y_i)
\right],
\end{equation}
where $w(t)$ emphasizes low-noise steps and $\ell_{\mathrm{bin}}$ is a focal classification loss. At inference, we threshold $\sigma(s_i)$ to discard tokens that the model classifies as absent before decoding the remaining latents with the VAE decoder.

\subsubsection{Aerial Conditioning.}
Given an aerial image $\mathcal{I}$, we extract patch tokens using a frozen DINOv3~\cite{simeoni_dinov3_2025} ViT-S16+ backbone $f_\mathrm{DINO}$, selected for its ability to produce semantically rich representations robust to environmental variations. Following the findings of Siméoni~\etal for dense feature extraction, we use a set of intermediate transformer blocks $L_{\mathrm{DINO}} = \{2,5,8,11\}$ and fuse their patch features into a single representation~\cite{simeoni_dinov3_2025}.
To preserve correspondence to the image grid, we add a $2\mathrm{D}$ sine-cosine positional encoding $\mathbf{E}$ with a learnable scale $\lambda_{\mathrm{pe}}$.
A transformer-based, trainable aerial token adapter $g_{\theta}$ then refines the fused tokens before they are provided to the denoiser
\begin{equation}
\mathbf{U}^{(k)} = f_{\mathrm{DINO}}^{(k)}(\mathcal{I}), 
\quad
\mathbf{C} = g_{\theta}\!\left(\mathrm{Concat}_{k\in L_{\mathrm{DINO}}}\mathbf{U}^{(k)} + \lambda_{\mathrm{pe}}\mathbf{E}\right).
\end{equation}
Rather than collapsing this information into a single global context vector, we inject the token sequence $\mathbf{C}$ through multi-head cross-attention within the DiT blocks.
Conditioning on local image features constrains the search space, shifting the objective from generating any realistic map to reconstructing the specific map depicted in the aerial view.
During training, we additionally apply conditioning dropout by zeroing the aerial token set for a subset of samples, enabling classifier-free guidance at inference.

\subsection{Global Map Generation}
\label{subsec:mapdreamer_global_map}
To scale the local, tile-based inference to city-scale mapping over a global area $\Omega_{\mathrm{g}}$, we discretize $\Omega_{\mathrm{g}}$ into a grid of overlapping tiles $\Omega_{i,j}$ and run inference in a fixed raster order from bottom-left to top-right as illustrated in~\cref{fig:global_graph_inference}.
Rowe \etal employ scene inpainting to extend generation beyond tile boundaries~\cite{rowe_scenario_2025}. However, we found this approach to be sensitive to error accumulation for map generation, since predictions in the overlap become conditioning for subsequent tiles, which can introduce false-positive lanes and suppress false negative lanes.
We instead propose \textit{boundary attention}, which injects compact boundary features from already predicted neighboring tiles through a dedicated cross-attention branch into the denoiser for each DiT block.
For the current tile $\Omega_{i,j}$, we extract boundary intersection features from the accumulated graph and embed them as boundary tokens $\{\mathbf{b}_k \}_{k=1}^{N_\mathrm{b}^{(i,j)}}$, with
\begin{equation}
     \mathbf{b}_k = f_\mathrm{c}(\left[x_k,y_k,\delta x_k, \delta y_k, \mathbf{s}_k\right])  \in \mathbb{R}^{d_\mathrm{b}},
\end{equation}
where $(x_k,y_k)$ denotes the intersection point with the tile boundary, $(\delta x_k,\delta y_k)$ the local tangent direction, and $\mathbf{s}_k$ is a one-hot indicator of the boundary side. These features are visualized in~\cref{fig:global_graph_inference}.
We identify two distinct boundary conditions that occur in our rasterized generation order from bottom-left to top-right.
During training, with some probability, we keep only one neighbor's context by dropping either the left or bottom boundary token set, simulating tiles with a single previously generated neighbor.
The second case occurs when a tile has two neighbors whose lane graphs have already been predicted. In this case, the context forms an L-shaped region comprising the left and bottom neighbors.

\section{Experiments}
\label{sec:experiments}
\subsection{Data}
\label{sec:data}
We build MapDreamer on Argoverse 2~\cite{wilson_argoverse_2023} map data with aligned aerial imagery from~\cite{buchner_learning_2023}, following the UrbanLaneGraph geospatial split for the training and validation regions.
We curate training tiles from the Argoverse 2 Motion Forecasting (AV2-MF)~\cite{wilson_argoverse_2023} samples and filter spatial redundancy by enforcing a maximum IoU of 0.8 for training and 0.4 for validation, resulting in $\approx 82{,}000$ training samples and $\approx 1{,}200$ geospatially strictly separated validation samples. To improve image-to-map consistency, we refine the UrbanLaneGraph alignment by estimating rigid transformations on a finer grid, specifically $2\,\mathrm{km}\times2\,\mathrm{km}$ map chunks. 

During training, we apply random rotations and translations to the aerial image and its aligned geometry. Since crops near tile boundaries can introduce artificial road terminations from missing ground truth, we use topology-aware rejection sampling to discard augmentation parameters that would move these boundary truncations into view.

\subsection{Metrics}
\label{subsec:metrics}
Both the geometric and topological correctness of the generated map are essential for autonomous driving functions. 
We therefore follow the metrics used by the state-of-the-art works BGFormer~\cite{blayney_bezier_2024} and LaneGNN~\cite{buchner_learning_2023}.

\textbf{GEO} is designed to evaluate the spatial accuracy of the generated map.
We adapt the metric implementation in BGFormer~\cite{blayney_bezier_2024} and discretize all lane centerlines into dense point sets of four nodes per meter to approximate continuous curves.
Unlike simple nearest-neighbor approaches, the evaluation is formulated as a one-to-one assignment problem. We compute the optimal bipartite matching between the predicted point set and the ground truth set that minimizes the total Euclidean distance.
We constrain the assignment by spatial proximity through a distance threshold of $\omega$, and by heading, requiring the tangent vectors of matched points to align within $\theta=\pi/3$, to enforce correct lane directionality between matched points.
Matches satisfying both conditions are considered \textit{true positives}, we report Precision, Recall, and F1-score for GEO.

\textbf{TOPO}, originally introduced by Biagioni and Eriksson~\cite{biagioni_inferring_2012}, jointly assesses geometric accuracy and topological connectivity using local subgraph comparison. We adopt the BGFormer~\cite{blayney_bezier_2024} implementation and propose a topology-aware sampling strategy to replace dense uniform sampling.
Instead of evaluating every matched point obtained through the matching principles of GEO, we define a set of critical starting locations derived from the ground truth topology to focus on complex topologies and avoid evaluating irrelevant graph sections. 
We target edges connected to nodes with $\mathrm{deg}(v)>2$. For lane splits, we initialize sampling $10\,\mathrm{m}$ upstream of the junction and traverse a total of $50\,\mathrm{m}$ downstream. For merging lanes, we apply the reverse traversal.
Additionally, to maintain coverage of simple geometries, we include starting points on all edges with $\mathrm{deg}(v)=1$ and within any straight section exceeding $20\,\mathrm{m}$.
All sampled subgraphs are then evaluated using GEO, thereby measuring lane-level similarity of complex topologies to the ground truth.

\textbf{Rasterized Intersection over Union} is robust to vertex ordering and point density in the lane graph and therefore serves as a visual metric. Predicted and ground truth lanes are rasterized into binary images with a fixed line width of $2\,\omega$, and their Intersection over Union (IoU) is computed. 

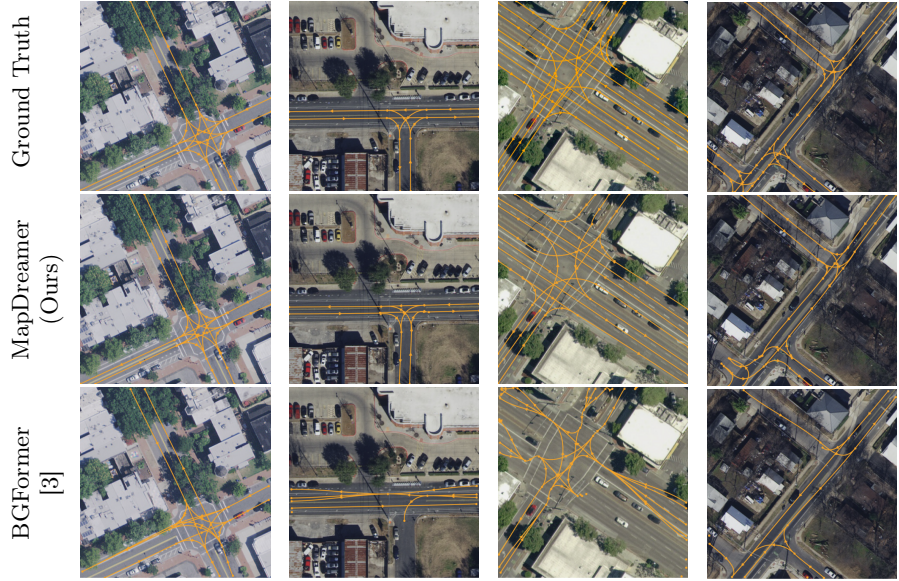
\begin{figure}[t]
    \centering
    \def\svgwidth{\linewidth} 
\begingroup%
  \makeatletter%
  \providecommand\color[2][]{%
    \errmessage{(Inkscape) Color is used for the text in Inkscape, but the package 'color.sty' is not loaded}%
    \renewcommand\color[2][]{}%
  }%
  \providecommand\transparent[1]{%
    \errmessage{(Inkscape) Transparency is used (non-zero) for the text in Inkscape, but the package 'transparent.sty' is not loaded}%
    \renewcommand\transparent[1]{}%
  }%
  \providecommand\rotatebox[2]{#2}%
  \newcommand*\fsize{\dimexpr\f@size pt\relax}%
  \newcommand*\lineheight[1]{\fontsize{\fsize}{#1\fsize}\selectfont}%
  \ifx\svgwidth\undefined%
    \setlength{\unitlength}{345.82677165bp}%
    \ifx\svgscale\undefined%
      \relax%
    \else%
      \setlength{\unitlength}{\unitlength * \real{\svgscale}}%
    \fi%
  \else%
    \setlength{\unitlength}{\svgwidth}%
  \fi%
  \global\let\svgwidth\undefined%
  \global\let\svgscale\undefined%
  \makeatother%
  \begin{picture}(1,0.62704918)%
    \lineheight{1}%
    \setlength\tabcolsep{0pt}%
    \put(0.04587464,0.44643524){\rotatebox{90}{\makebox(0,0)[lt]{\lineheight{1.20000005}\smash{\begin{tabular}[t]{l}Ground Truth\end{tabular}}}}}%
    \put(0.0452674,0.31513226){\rotatebox{90}{\makebox(0,0)[t]{\lineheight{1.20000005}\smash{\begin{tabular}[t]{c}MapDreamer\\(Ours)\end{tabular}}}}}%
    \put(0.04567223,0.1058045){\rotatebox{90}{\makebox(0,0)[t]{\lineheight{1.20000005}\smash{\begin{tabular}[t]{c}BGFormer\\\cite{blayney_bezier_2024}\end{tabular}}}}}%
    \put(0,0){\includegraphics[width=\unitlength,page=1]{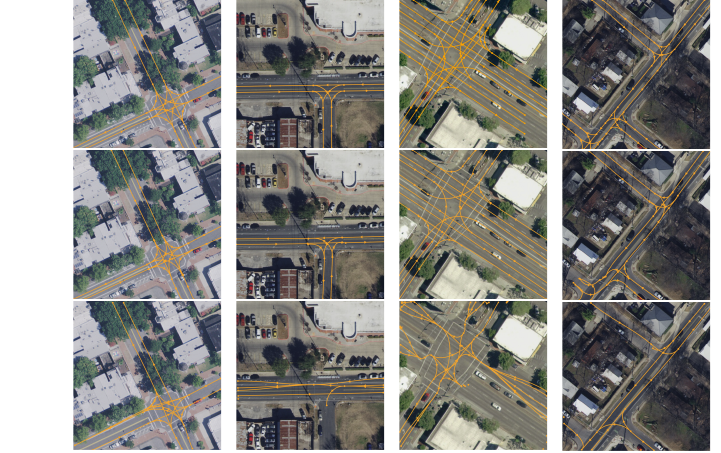}}%
  \end{picture}%
\endgroup%

    \caption{Qualitative results for local lane graph prediction from aerial imagery across multiple cities, comparing MapDreamer with BGFormer~\cite{blayney_bezier_2024}.}
    \label{fig:results:qualitative_local}
\end{figure}

\subsection{Results}
\label{subsec:results}
We compare MapDreamer with the state-of-the-art models in lane graph prediction from aerial imagery. To be able to evaluate the models on the adopted metrics, we retrain BGFormer~\cite{blayney_bezier_2024}, reporting the best results of three runs. For evaluation of LaneGNN~\cite{buchner_learning_2023}, we use the provided checkpoints. 
LaneGNN~\cite{buchner_learning_2023} initializes graph traversal from $178$ starting poses produced by LaneExtraction~\cite{he_lane-level_2022}. Since we could not reproduce this, we initialize the method using $178$ poses sampled from the ground truth.
We report the metrics for two matching radii: a strict threshold ($\omega_{1}=1.2\,\mathrm{m}$) and a relaxed threshold ($\omega_{2}=2\,\mathrm{m}$).

\subsubsection{Local Lane Graph Evaluation.}
LaneGNN~\cite{buchner_learning_2023} does not support local tile graph evaluation on our data since it requires a lane starting at the bottom center of the crop. We therefore compare MapDreamer to BGFormer~\cite{blayney_bezier_2024}, evaluating both methods on the same validation split of our dataset. For local lane graph generation, the mean over all samples is reported.

\Cref{tab:results:qualitative_local_global} shows that MapDreamer improves both geometric accuracy and topological connectivity substantially compared to BGFormer~\cite{blayney_bezier_2024} under both tolerances. For the strict setting $\omega_1 = 1.2\,\mathrm{m}$, MapDreamer increases $\mathrm{GEO}_1$ by $\fpeval{round(100*(\MDGEOsmallF-\BGFGEOsmallF),2)}\%$ in $\mathrm{F}_1$ and raises the rasterized $\mathrm{IoU}_1$ from $\BGFIoUsmall$ to $\MDIoUsmall$.
The $\mathrm{TOPO}_1$ precision and recall also improve substantially, by $\fpeval{round(100*(\MDTOPOsmallP-\BGFTOPOsmallP),2)}\%$ and $\fpeval{round(100*(\MDTOPOsmallR-\BGFTOPOsmallR),2)}\%$ respectively, indicating fewer locally inconsistent junction traversals at the split- and merge-focused starting locations used by our TOPO sampling. For the relaxed setting $\omega_2 = 2\,\mathrm{m}$, the same trend holds, with $\mathrm{GEO}_2$ $\mathrm{F}_1$ improving by $\fpeval{round(100*(\MDGEObigF-\BGFGEObigF),2)}\%$ 
and $\mathrm{TOPO}_2$ $\mathrm{F}_1$ improving by $\fpeval{round(100*(\MDTOPObigF-\BGFTOPObigF),2)}\%$, 
suggesting that the gain is not only driven by small offsets but also by better structural reconstruction.
The quantitative results align with the qualitative results shown in \cref{fig:results:qualitative_local}. 
BGFormer fails abruptly when confidence thresholding removes nodes or edges due to incorrect cardinality estimates, whereas MapDreamer degrades more gracefully and remains stable under such count and connectivity errors, see columns two and three in \cref{fig:results:qualitative_local}.
\begin{table}[t]
\centering
\caption{Quantitative results for the local tile-based evaluation and the global lane graph generation for both matching thresholds $\omega_1=1.2\,\mathrm{m}$ and $\omega_2=2\,\mathrm{m}$. See~\cref{subsec:metrics} for a description of the metrics. (P/R/$\mathrm{F}_1$) denote precision, recall and $\mathrm{F}_1$ score, IoU the rasterized Intersection over Union. 
Note that LaneGNN is initialized using 178 ground-truth poses for each global tile.}
\label{tab:results:qualitative_local_global}
\resizebox{\textwidth}{!}{%
\begin{tabular}{@{}lccclccccccclcccc@{}}
\toprule
\multirow{2}{*}{Method} &
  \multicolumn{3}{c}{$\mathrm{GEO}_1$} &
  \multirow{2}{*}{} &
  \multicolumn{3}{c}{$\mathrm{TOPO}_1$} &
  \multirow{2}{*}{$\mathrm{IoU}_1$} &
  \multicolumn{3}{c}{$\mathrm{GEO}_2$} &
   &
  \multicolumn{3}{c}{$\mathrm{TOPO}_2$} &
  \multirow{2}{*}{$\mathrm{IoU}_2$} \\ \cmidrule(lr){2-4} \cmidrule(lr){6-8} \cmidrule(lr){10-12} \cmidrule(lr){14-16}
 &
  $\mathrm{P}$ &
  $\mathrm{R}$ &
  $\mathrm{F}_1$ &
   &
  $\mathrm{P}$ &
  $\mathrm{R}$ &
  $\mathrm{F}_1$ &
   &
  $\mathrm{P}$ &
  $\mathrm{R}$ &
  $\mathrm{F}_1$ &
   &
  $\mathrm{P}$ &
  $\mathrm{R}$ &
  $\mathrm{F}_1$ &
   \\ \midrule
\multicolumn{17}{c}{Local Lane Graph} \\ \midrule
BGFormer~\cite{blayney_bezier_2024} &
  \BGFGEOsmallP &
  \BGFGEOsmallR &
  \BGFGEOsmallF &
   &
  \BGFTOPOsmallP &
  \BGFTOPOsmallR &
  \BGFTOPOsmallF &
  \BGFIoUsmall &
  \BGFGEObigP &
  \multicolumn{1}{l}{\BGFGEObigR} &
  \multicolumn{1}{l}{\BGFGEObigF} &
   &
  \BGFTOPObigP &
  \multicolumn{1}{l}{\BGFTOPObigR} &
  \multicolumn{1}{l}{\BGFTOPObigF} &
  \BGFIoUbig \\
MapDreamer (Ours) &
  \textbf{\MDGEOsmallP} &
  \textbf{\MDGEOsmallR} &
  \textbf{\MDGEOsmallF} &
   &
  \textbf{\MDTOPOsmallP} &
  \textbf{\MDTOPOsmallR} &
  \textbf{\MDTOPOsmallF} &
  \textbf{\MDIoUsmall} &
  \textbf{\MDGEObigP} &
  \multicolumn{1}{l}{\textbf{\MDGEObigR}} &
  \multicolumn{1}{l}{\textbf{\MDGEObigF}} &
   &
  \textbf{\MDTOPObigP} &
  \multicolumn{1}{l}{\textbf{\MDTOPObigR}} &
  \multicolumn{1}{l}{\textbf{\MDTOPObigF}} &
  \textbf{\MDIoUbig} \\ \midrule
\multicolumn{17}{c}{Global Lane Graph} \\ \midrule
LaneGNN~\cite{buchner_learning_2023} &
  \textbf{\LGNNGEOsmallPglobal} &
  \LGNNGEOsmallRglobal &
  \LGNNGEOsmallFglobal &
   &
  \LGNNTOPOsmallPglobal &
  \LGNNTOPOsmallRglobal &
  \LGNNTOPOsmallFglobal &
  \LGNNIoUsmallglobal &
  \textbf{\LGNNGEObigPglobal} &
  \LGNNGEObigRglobal &
  \LGNNGEObigFglobal &
   &
  \LGNNTOPObigPglobal &
  \LGNNTOPObigRglobal &
  \LGNNTOPObigFglobal &
  \LGNNIoUbigglobal \\
BGFormer~\cite{blayney_bezier_2024} &
  \BGFGEOsmallPglobal &
  \BGFGEOsmallRglobal &
  \BGFGEOsmallFglobal &
   &
  \BGFTOPOsmallPglobal &
  \BGFTOPOsmallRglobal &
  \BGFTOPOsmallFglobal &
  \BGFIoUsmallglobal &
  \BGFGEObigPglobal &
  \BGFGEObigRglobal &
  \BGFGEObigFglobal &
   &
  \BGFTOPObigPglobal &
  \BGFTOPObigRglobal &
  \BGFTOPObigFglobal &
  \BGFIoUbigglobal \\
MapDreamer (Ours) &
  \MDGEOsmallPglobal &
  \textbf{\MDGEOsmallRglobal} &
  \textbf{\MDGEOsmallFglobal} &
   &
  \textbf{\MDTOPOsmallPglobal} &
  \textbf{\MDTOPOsmallRglobal} &
  \textbf{\MDTOPOsmallFglobal} &
  \textbf{\MDIoUsmallglobal} &
  \MDGEObigPglobal &
  \textbf{\MDGEObigRglobal} &
  \textbf{\MDGEObigFglobal} &
   &
  \textbf{\MDTOPObigPglobal} &
  \textbf{\MDTOPObigRglobal} &
  \textbf{\MDTOPObigFglobal} &
  \textbf{\MDIoUbigglobal} \\ \bottomrule
\end{tabular}%
}
\end{table}
\subsubsection{Global Lane Graph Evaluation.}
\begin{figure}
    \centering 
    \begin{subfigure}{0.39\linewidth}
        \def\svgwidth{1.0\linewidth} 
        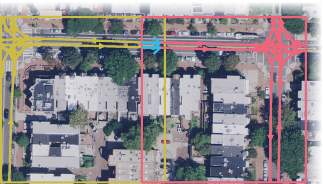
        \caption{Visualization of the global graph inference strategy for two neighboring tiles $\Omega_{i,j}$ and $\Omega_{i+1,j}$. The boundary features $\{\mathbf{b}_k\}_{k=1}^{2}$ for $\Omega_{i+1,j}$ are visualized using blue arrows.}      
        \label{fig:global_graph_inference}
    \end{subfigure}
    \hfill
    \begin{subfigure}{0.59\linewidth}
        \centering
        \def\svgwidth{\linewidth} 
        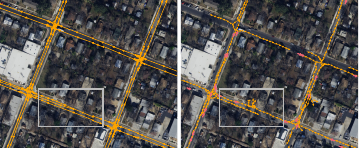
        \caption{Qualitative results of MapDreamer (left) and LaneGNN~\cite{buchner_learning_2023} (right) on the global lane graph tasks. For LaneGNN, pink arrows indicate graph traversal initialization points sampled from ground truth. The boxed area highlights a challenging lane split.}
        \label{fig:results:qualitative_global}    
    \end{subfigure}
    \caption{Global graph inference strategy using boundary features (\subref{fig:global_graph_inference}) and qualitative results for global graph generation (\subref{fig:results:qualitative_global}).}
    \label{fig:global_graph}
\end{figure}

We evaluate MapDreamer on the global $5000\,\mathrm{px} \times 5000\,\mathrm{px}$ evaluation tiles from the UrbanLaneGraph dataset~\cite{buchner_learning_2023}. 
During global inference, we use a $32\,\mathrm{px}$ rejection margin on all subtile sides $\Omega_{i,j}$ to suppress noisy boundary artifacts.
We compare our method against the state-of-the-art baseline methods LaneGNN~\cite{buchner_learning_2023} and BGFormer~\cite{blayney_bezier_2024}, 
\cref{tab:results:qualitative_local_global} summarizes the results.
MapDreamer retains a clear advantage over the Bézier-curve based BGFormer~\cite{blayney_bezier_2024}, improving $\mathrm{GEO}_1$ by $\fpeval{round(100*(\MDGEOsmallFglobal-\BGFGEOsmallFglobal),2)}\%$ and $\mathrm{TOPO}_1$ by $\fpeval{round(100*(\MDTOPOsmallFglobal-\BGFTOPOsmallFglobal),2)}\%$ in $\mathrm{F}_1$.
Compared to LaneGNN, MapDreamer attains higher recall and more balanced precision-recall trade-offs, achieving an improvement of $\mathrm{GEO}_1$ by $\fpeval{round(100*(\MDGEOsmallFglobal-\LGNNGEOsmallFglobal),2)}\%$ and $\mathrm{TOPO}_1$ by $\fpeval{round(100*(\MDTOPOsmallFglobal-\LGNNTOPOsmallFglobal),2)}\%$ in $\mathrm{F}_1$. \Cref{fig:results:qualitative_global} shows qualitative results on a global evaluation tile crop for MapDreamer and LaneGNN, highlighting MapDreamer's ability to correctly predict geometry and topology even in challenging regions. 
We report complete evaluation tiles for different cities in the supplementary material.

\subsubsection{Failure Cases.}
\begin{figure}[t]
    \centering 
    \begin{subfigure}{0.24\linewidth}
        \includegraphics[width=\linewidth]{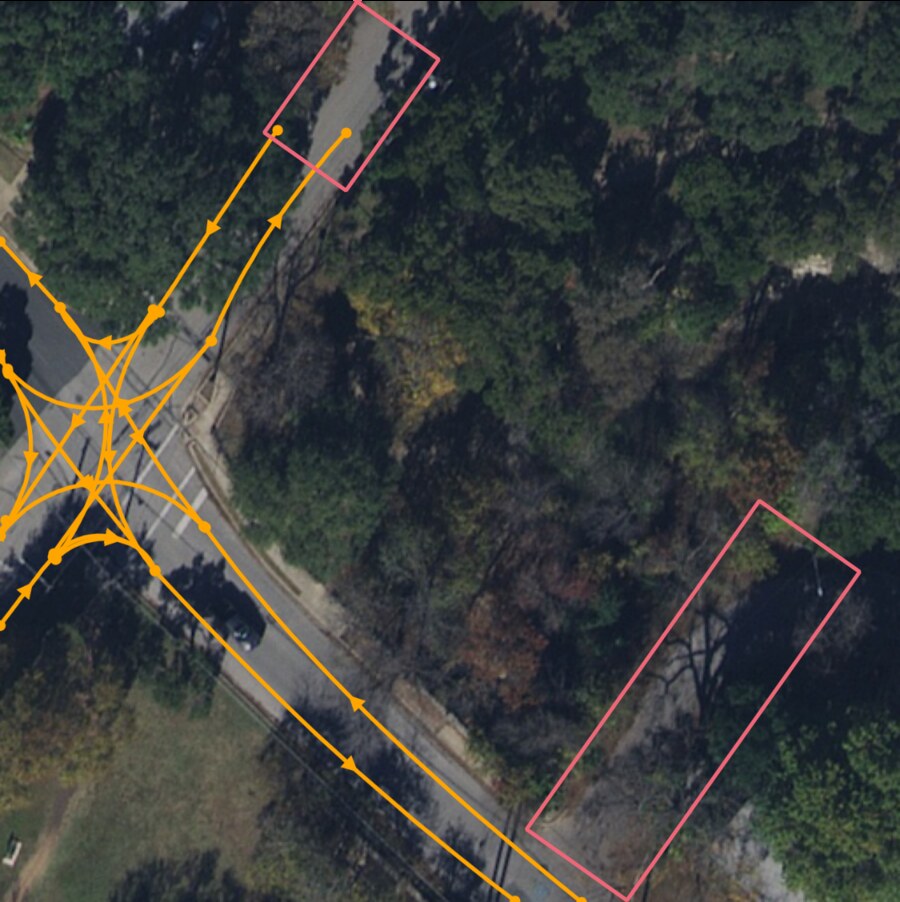}
        \caption{Occlusions causing under-prediction.}      
        \label{fig:failure_case_1}
    \end{subfigure}
    \hfill
    \begin{subfigure}{0.24\linewidth}
        \includegraphics[width=\linewidth]{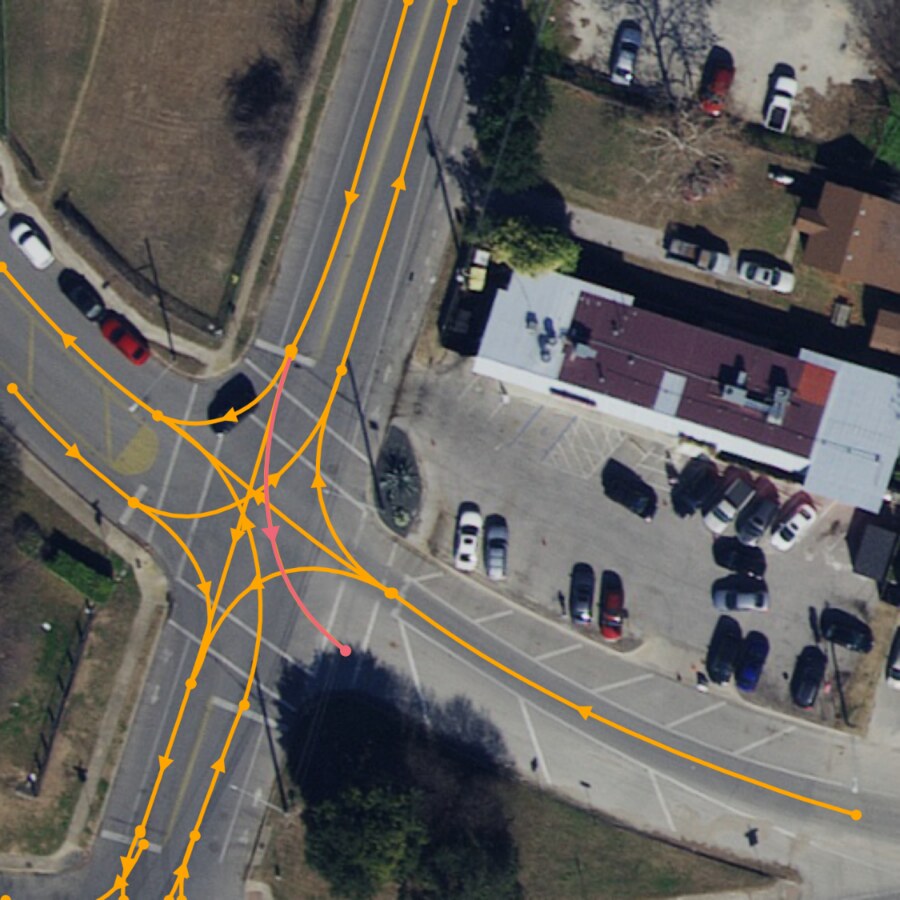}
        \caption{Incorrect turn toward one-way road.}      
        \label{fig:failure_case_2}
    \end{subfigure}
    \hfill
    \begin{subfigure}{0.24\linewidth}
        \includegraphics[width=\linewidth]{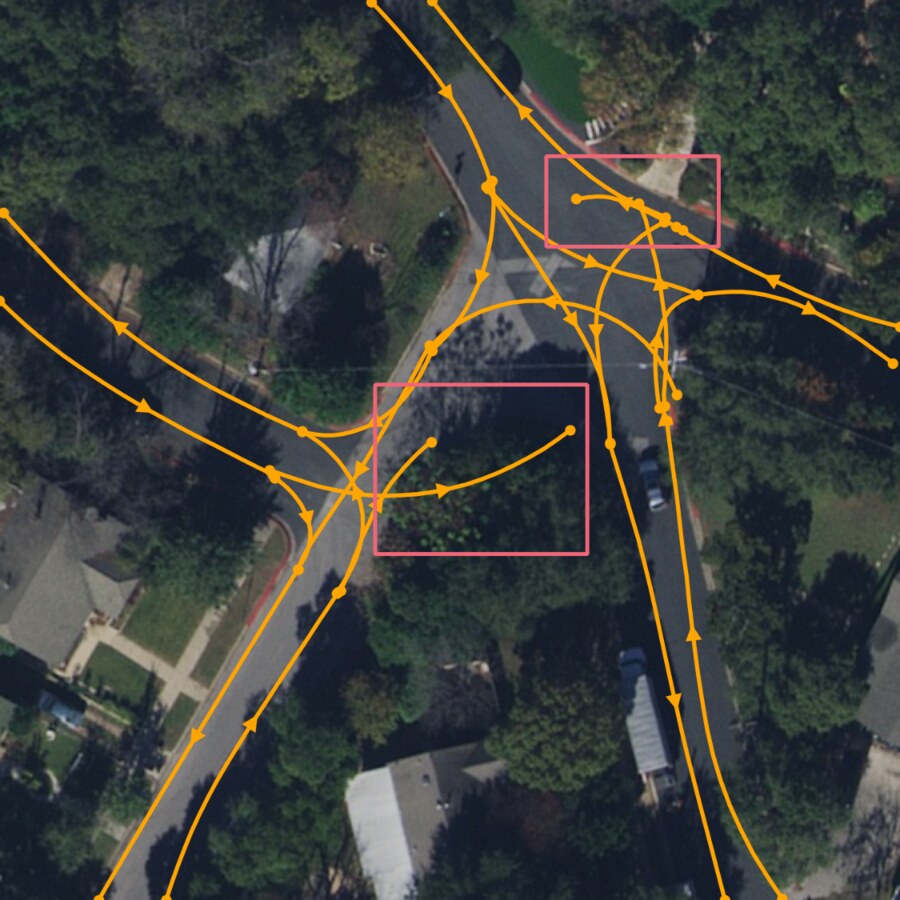}
        \caption{Complex intersection w/o lane markings.}      
        \label{fig:failure_case_3}
    \end{subfigure}    
    \hfill
    \begin{subfigure}{0.24\linewidth}
        \includegraphics[width=\linewidth]{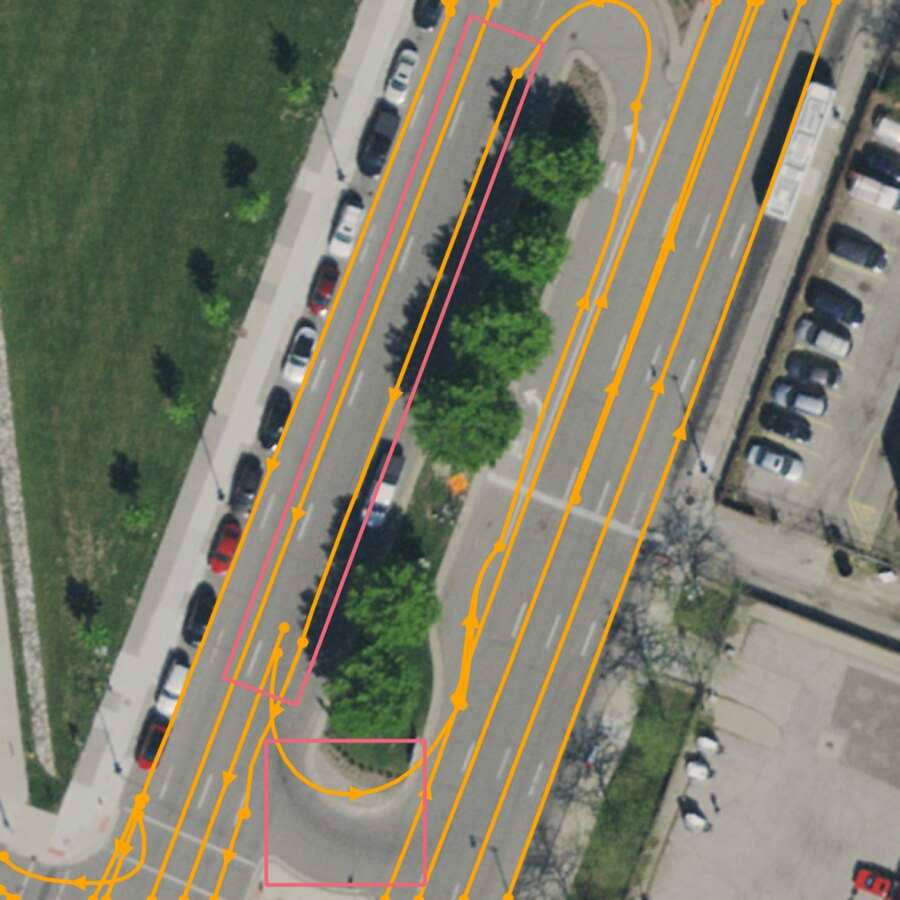}
        \caption{Missing lane and shifted geometry.}      
        \label{fig:failure_case_4}
    \end{subfigure}    
    \caption{Local lane graph failure cases of MapDreamer visualized using pink boxes for missing predictions and geometry errors, and pink lanes for wrong predictions.}
    \label{fig:failure_modes}
\end{figure}

\Cref{fig:failure_modes} shows representative failure modes of MapDreamer.
Errors occur in areas with limited visual evidence, such as heavy occlusions from trees or missing lane markings, which cause under-prediction of occluded lanes (\cref{fig:failure_modes}\subref{fig:failure_case_1}, \subref{fig:failure_case_3}) and over-prediction of plausible but unsupported connectivity (\cref{fig:failure_modes}\subref{fig:failure_case_2}).
Rare or underrepresented configurations, such as unusual intersections without lane markings (\cref{fig:failure_modes}\subref{fig:failure_case_3}), or U-turns (\cref{fig:failure_modes}\subref{fig:failure_case_4}), remain challenging but could likely be mitigated in future work with more diverse training data.

\subsection{Ablations}
\label{subsec:ablations}
We study the contribution of the lane cardinality module, ghost lane latents and the latent space encoding through the VAE. \cref{tab:results:ablations} summarizes the results of our ablations for the strict threshold $\omega_1=1.2\,\mathrm{m}$. We provide qualitative results for the ablations in the supplementary material. All configurations using the VAE are trained on the same VAE checkpoint.
\begin{table}[b]
    \centering
    \caption{Results of the ablation studies on the integration of the Lane Cardinality module (LC), Ghost Lane Latents (GL), and the latent diffusion in the Variational Autoencoder (VAE) latent space for the local tile-based evaluation.
    }
    \resizebox{0.6\textwidth}{!}{
    \label{tab:results:ablations}
    \begin{tabular}{@{}cccccccccc@{}}
        \toprule
        \multirow{2}{*}{LC} &
          \multirow{2}{*}{GL} &
          \multirow{2}{*}{VAE} &
          \multicolumn{3}{c}{$\mathrm{GEO}_1$} &
          \multicolumn{3}{c}{$\mathrm{TOPO}_1$} &
          \multirow{2}{*}{$\mathrm{IoU}_1$} \\
         &
           &
           &
          $\mathrm{P}$ &
          $\mathrm{R}$ &
          $\mathrm{F}_1$ &
          $\mathrm{P}$ &
          $\mathrm{R}$ &
          $\mathrm{F}_1$ &
           \\ \midrule
        \checkmark &
          \checkmark &
          \checkmark &
          $\mathbf{\MDGEOsmallP}$ &
          $\mathbf{\MDGEOsmallR}$ &
          $\mathbf{\MDGEOsmallF}$ &
          $\mathbf{\MDTOPOsmallP}$ &
          $\mathbf{\MDTOPOsmallR}$ &
          $\mathbf{\MDTOPOsmallF}$ &
          $\mathbf{\MDIoUsmall}$ \\
        \checkmark &
          $\times$ &
          \checkmark &
          \MDGEOsmallPLCnoGL &
          \MDGEOsmallRLCnoGL &
          \MDGEOsmallFLCnoGL &
          \MDTOPOsmallPLCnoGL &
          \MDTOPOsmallRLCnoGL &
          \MDTOPOsmallFLCnoGL &
          \MDIoUsmallLCnoGL \\
        $\times$ &
          $\times$ &
          \checkmark &
          \MDGEOsmallPnoLCnoGL &
          \MDGEOsmallRnoLCnoGL &
          \MDGEOsmallFnoLCnoGL &
          \MDTOPOsmallPnoLCnoGL &
          \MDTOPOsmallRnoLCnoGL &
          \MDTOPOsmallFnoLCnoGL &
          \MDIoUsmallnoLCnoGL \\ \midrule
        \checkmark &
          $\times$ &
          $\times$ &
          \MDGEOsmallPnoVAE &
          \MDGEOsmallRnoVAE &
          \MDGEOsmallFnoVAE &
          \MDTOPOsmallPnoVAE &
          \MDTOPOsmallRnoVAE &
          \MDTOPOsmallFnoVAE &
          \MDIoUsmallnoVAE \\ \bottomrule
    \end{tabular}
    }%
\end{table}
\subsubsection{Ghost Lane Latents and Lane Cardinality.}
For the first ablation, we keep the lane cardinality module and the diffusion pipeline unchanged, but disable the appended ghost lane latent slots, which keeps the lane count estimate $\tilde{N}_\mathrm{l}$ as the sole mechanism to set the number of active lane slots. As a result, $\mathrm{GEO}_1$ precision drops by $\fpeval{round(100*(\MDGEOsmallP-\MDGEOsmallPLCnoGL),2)}\%$ while the recall remains nearly unchanged, which points to more spurious or duplicated lane instances rather than systematic omission of centerlines. In contrast, $\mathrm{TOPO}_1$ recall decreases more strongly by $\fpeval{round(100*(\MDTOPOsmallR-\MDTOPOsmallRLCnoGL),2)}\%$, consistent with underestimation of leaving some lanes uninstantiated and thereby removing multiple adjacency relations. Overall, ghost latents improve slot allocation under count mismatch, which stabilizes connectivity and reduces false positives.

In the fixed slot setting, we remove the lane cardinality module and ghost latents and instead pad each sample to a constant number $N_\mathrm{l}^\mathrm{fs}=64$ of lane tokens from Gaussian latent noise. During training, the geometric diffusion loss is applied only on existing lanes, while the existence head is supervised on all slots.
While $\mathrm{GEO}_1$ recall remains relatively consistent (-$\fpeval{round(100*(\MDGEOsmallR-\MDGEOsmallRnoLCnoGL),2)}\%$), precision collapses by $\fpeval{round(100*(\MDGEOsmallP-\MDGEOsmallPnoLCnoGL),2)}\%$. This indicates overgeneration of lanes, where \textit{padded} slots drift into plausible lane hypotheses and survive thresholding, which also degrades topology metrics.

\subsubsection{Vectorized Diffusion Model.}
To assess the contribution of the latent feature encoding, we conduct an ablation in which we remove the VAE stage and train the same diffusion model directly in normalized vector coordinate space. 
This modification eliminates the geometric and topological prior imposed by the VAE decoder and shifts the regularization entirely to the diffusion transformer, making the design conceptually closer to the online map generation architecture MapDiffusion~\cite{monninger_mapdiffusion_2025}, where diffusion is also applied directly to polyline queries without an intermediate latent compression.
Lane to lane relations are predicted from the denoised lane features using a topology head of similar size to the VAE decoder head.
Since this ablation disables the ghost lane latents but retains the lane cardinality module, we compare with the latent MapDreamer results in the second row of \Cref{tab:results:ablations}.
Removing the VAE degrades both geometric and topological quality, $\mathrm{GEO}_1$ $\mathrm{F}_1$ drops by $\fpeval{round(100*(\MDGEOsmallFLCnoGL-\MDGEOsmallFnoVAE),2)}\%$, 
$\mathrm{TOPO}_1$ $\mathrm{F}_1$ by $\fpeval{round(100*(\MDTOPOsmallFLCnoGL-\MDTOPOsmallFnoVAE),2)}\%$. This suggests that the VAE and the inherent latent space promote consistent connectivity and geometry prediction representing the distribution of lane geometries from the dataset.

\subsubsection{Model Efficiency.}
We report parameter counts and per-tile inference time in \cref{tab:model_efficiency}.
To separate performance from model scale, we train a small MapDreamer variant by reducing the LDM DiT hidden dimension from $2048$ to $512$. Despite using fewer trainable parameters, it still outperforms BGFormer in both $\mathrm{GEO}_1$ and $\mathrm{TOPO}_1$ $\mathrm{F}_1$. 
Further, we add an ablation on DDIM sampling~\cite{song_ddim_2021} without retraining. Reducing denoising from $100$ DDPM steps to $20$ DDIM steps substantially lowers runtime with only minor performance loss, the small DDIM model matches BGFormer's runtime.
Additional hyperparameters and implementation details are provided in the supplementary material.

\begin{table}
\centering
\caption{Number of trainable parameters used during inference along with the inference time per local tile image $(512\times512\,\mathrm{px})$ on a single V100 GPU. For LaneGNN, we evaluate on the full tile $(5000\times5000\,\mathrm{px})$ and divide the inference time by $(5000/512)^2$.}
\label{tab:model_efficiency}
\resizebox{.7\columnwidth}{!}{%
\begin{tabular}{@{}lcclccclc@{}}
\toprule
 & \multirow{2}{*}{LaneGNN} & \multirow{2}{*}{BGFormer} & \, & \multicolumn{4}{c}{MapDreamer (Ours)} & \multicolumn{1}{l}{\,} \\ \cmidrule(l){4-9} 
 &  &  &  &  & Small & \multicolumn{2}{c}{Default} & \multicolumn{1}{l}{\,} \\ \midrule
Number Param. & 187.2M & 39.9M & \, & \begin{tabular}[c]{@{}c@{}}LDM\\ VAE\end{tabular} & \begin{tabular}[c]{@{}c@{}}21.5M\\ 9.5M\end{tabular} & \multicolumn{2}{c}{\begin{tabular}[c]{@{}c@{}}230.5M\\ 9.5M\end{tabular}} & \, \\
Inference time [ms] & 48457\,ms & 33\,ms & \, & \begin{tabular}[c]{@{}c@{}}DDPM\\ DDIM\end{tabular} & \begin{tabular}[c]{@{}c@{}}113\,ms\\ 34\,ms\end{tabular} & \multicolumn{2}{c}{\begin{tabular}[c]{@{}c@{}}326\,ms\\ 73\,ms\end{tabular}} & \, \\
($\mathrm{GEO}_1 \mid \mathrm{TOPO}_1$) $\mathrm{F}_1$ & - & 0.46 $\mid$ 0.35 & \, & \begin{tabular}[c]{@{}c@{}}DDPM\\ DDIM\end{tabular} & \multicolumn{1}{l}{\begin{tabular}[c]{@{}l@{}}0.59 $\mid$ 0.51 \\ 0.58 $\mid$ 0.50\end{tabular}} & \multicolumn{2}{c}{\begin{tabular}[c]{@{}c@{}}0.64 $\mid$ 0.56\\ 0.62 $\mid$ 0.54\end{tabular}} & \, \\ \bottomrule
\end{tabular}%
}
\end{table}
\section{Conclusion}
\label{sec:conclusion}
We presented MapDreamer, a generative approach to lane-level map generation from aerial imagery. MapDreamer combines a variational autoencoder trained on vectorized lane geometry and topology with a latent diffusion denoising model, enabling the generation of maps matching the distribution of real-world lane graphs.
Experiments on UrbanLaneGraph~\cite{buchner_learning_2023}, derived from Argoverse 2~\cite{wilson_argoverse_2023}, demonstrate improved geometric accuracy and topological connectivity over strong non-generative baselines in both local and global evaluation settings.

For the future, we aim to extend MapDreamer in several directions. Although lane centerlines and their connectivity are the most relevant for autonomous driving functions, we plan to extend prediction to additional map features such as lane boundaries and pedestrian crossings.
With the increasing availability of drone-based data capture, high-resolution aerial imagery is becoming easier to collect and update. 
However, the integration of complementary sources such as vehicle trajectories remains a promising direction, particularly for enabling more frequent map updates and improving robustness in regions where aerial cues are weak.
Finally, flow matching is a promising alternative to DDIM- and DDPM-based generation as it may further reduce sampling cost while preserving the benefits of iterative generative refinement.

%
%
\bibliographystyle{splncs04}
\bibliography{main}

\clearpage
\appendix
\section*{Supplementary Material}
\addcontentsline{toc}{section}{Supplementary Material}

\setcounter{figure}{0}
\setcounter{table}{0}
\renewcommand{\thefigure}{S\arabic{figure}}
\renewcommand{\thetable}{S\arabic{table}}

In the supplementary material of our work \textit{MapDreamer: Aerial Imagery Conditioned Latent Diffusion for 
Lane-Level Map Generation} we present details on the extraction, alignment, and characteristics of the dataset along with an extended results section.
Specifically, we present results of the graph aggregation on complete global tiles for different cities and detailed quantitative results.
Finally, we provide further ablation studies of the MapDreamer architecture.

\section{Dataset}
\label{sec:supp_data}
Most publicly available datasets for map generation focus on online inference from perspective-view images. We therefore first provide an overview of datasets that include aerial imagery before discussing the curation pipeline used in this work.
\subsection{Available Datasets}
\label{subsec_data_available}
The public UrbanLaneGraph~\cite{buchner_learning_2023} and LaneExtraction~\cite{he_lane-level_2022} datasets best fit the requirements of lane-level map prediction from aerial imagery.
UrbanLaneGraph~\cite{buchner_learning_2023} combines the map data from the public Argoverse 2 dataset~\cite{wilson_argoverse_2023} with aligned aerial imagery tiles from the six cities Austin, Miami, Pittsburgh, Palo Alto, Detroit, and Washington DC. In total, UrbanLaneGraph contains $5223.3\,\mathrm{km}$ of lane annotations compared to the $398.6\,\mathrm{km}$ of the second dataset published in LaneExtraction~\cite{he_lane-level_2022}. This difference in scale motivates our choice of the UrbanLaneGraph~\cite{buchner_learning_2023} aerial imagery as the basis for our dataset.

\subsection{MapDreamer Dataset}
\label{subsec:data_curation}
We construct our dataset samples using map data and scenario locations from the Argoverse 2 Motion Forecasting (AV2-MF) dataset~\cite{wilson_argoverse_2023}.
This approach allows us to retrieve spatial context beyond the target crop of $512\times512$ pixels, which is required for the rotations and translations used in our data augmentation pipeline.
To minimize spatial redundancy and ensure diverse coverage, we filter the $250{,}000$ scenarios in AV2-MF based on geospatial overlap. We enforce a maximum Intersection-over-Union (IoU) threshold of 0.8 for training samples and 0.4 for validation samples.

\subsubsection{Map Alignment.}
\label{ssub:data_map_alignment}
Büchner~\etal~\cite{buchner_learning_2023} use a set of three points to align each city's aerial image with the respective local coordinate system in Argoverse 2 employing the Kabsch-Umeyama~\cite{umeyama_least-squares_1991} algorithm. We found this alignment to be imprecise for parts of the map. For data curation, we therefore split the map data into tiles of $2\,\mathrm{km}\times2\,\mathrm{km}$ and aligned each tile individually, which improved the alignment substantially. 

\section{Extended Results}
Due to space limitations, the main paper contains only selected qualitative results.
In the following, we extend the results section with a more detailed quantitative analysis of global graph aggregation and additional qualitative examples, including complete global aerial image tiles. We further provide qualitative visualizations for the ablations presented in the main paper.

\subsection{Global Lane Graph Evaluation}
\subsubsection{Quantitative Results.}
\begin{figure}
    \centering
    \begin{subfigure}[b]{0.49\linewidth}
        \centering
        \def\svgwidth{\linewidth} 
        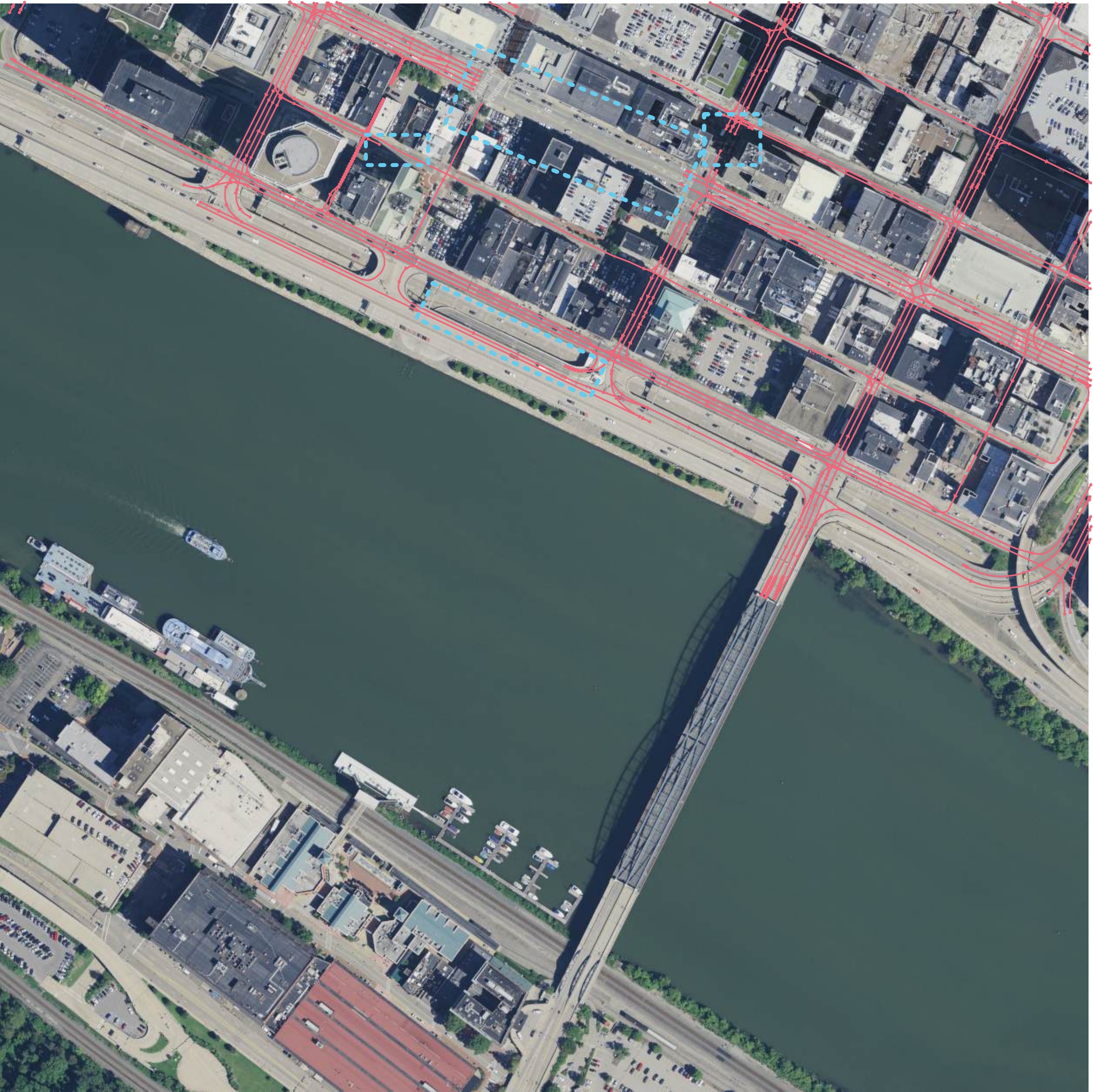
        \caption{Pittsburgh ID 5}
        \label{fig:pittsburgh_5_gt}
    \end{subfigure}
    \begin{subfigure}[b]{0.49\linewidth}
        \centering
        \def\svgwidth{\linewidth} 
        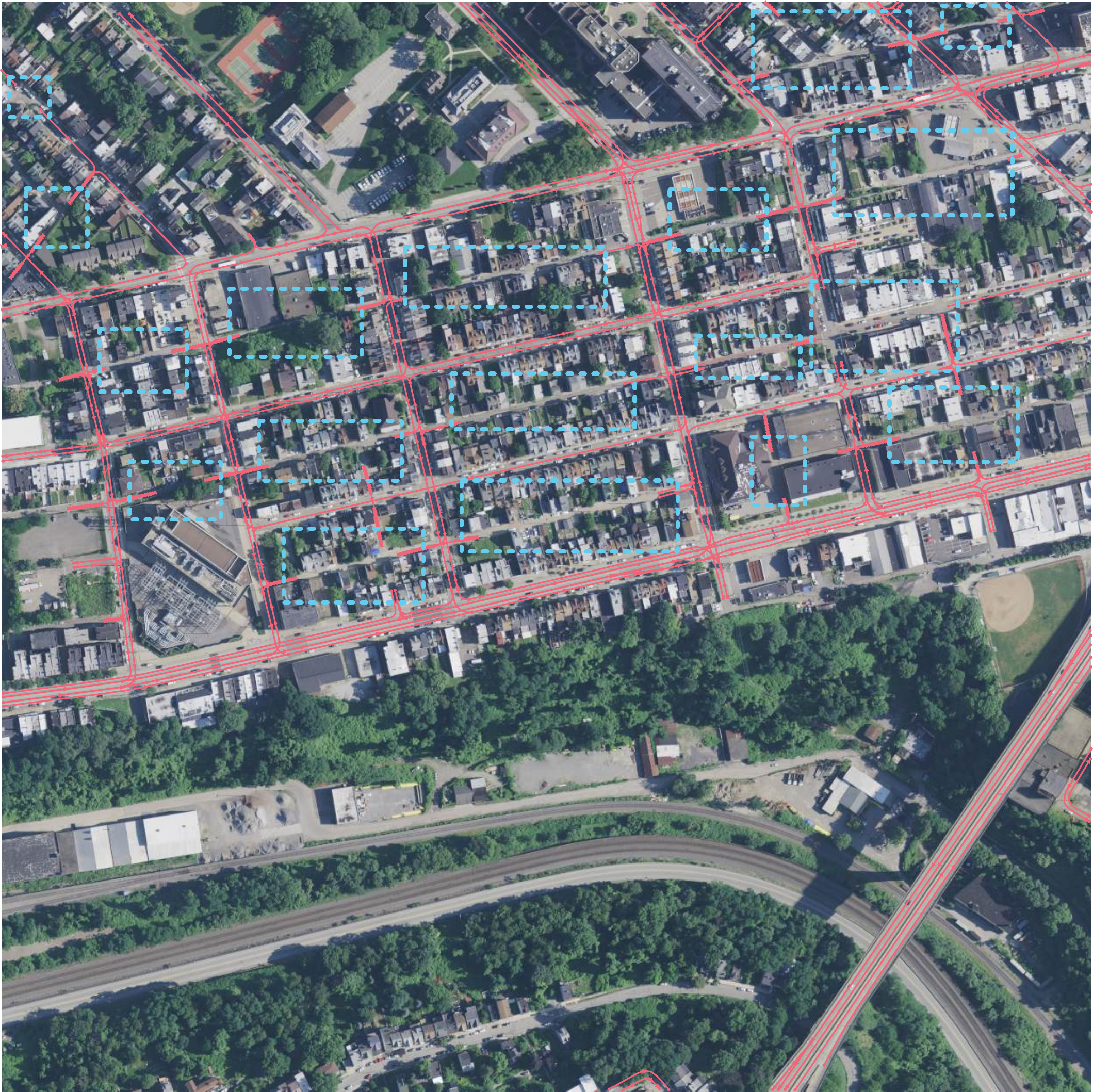
        \caption{Pittsburgh ID 36}
        \label{fig:pittsburgh_36_gt}
    \end{subfigure}
    \caption{Pittsburgh evaluation tiles from the UrbanLaneGraph~\cite{buchner_learning_2023} dataset. We mark ground truth issues affecting the correct evaluation of metrics, specifically the TOPO metric, which is particularly sensitive to road continuity, with \textcolor[HTML]{66CCEE}{cyan} dotted bounding boxes.}
    \label{fig:pittsburgh_gt_issues}
\end{figure}

The UrbanLaneGraph~\cite{buchner_learning_2023} dataset contains a total of eleven $5000\,\mathrm{px}\times 5000\,\mathrm{px}$ evaluation tiles with ground truth data accumulated from Argoverse 2~\cite{wilson_argoverse_2023} scenarios.
\Cref{tab:quantitative_global_per_tile} reports the quantitative per-tile results for the matching threshold $\omega_1=1.2\,\mathrm{m}$. 
We observe substantially lower scores on the Pittsburgh tiles across all three methods, particularly for $\mathrm{TOPO}_1$.
Visual inspection revealed issues in the accumulated ground truth, as illustrated in \cref{fig:pittsburgh_gt_issues}. We therefore exclude the Pittsburgh tiles from the reported mean values in the main paper but report them in the supplementary material.

\begin{table}
\centering
\caption{Quantitative results for the global lane graph generation on the strict matching threshold $\omega_1=1.2\,\mathrm{m}$ for each of the eleven $5000\,\mathrm{px}\times5000\,\mathrm{px}$ evaluation tiles from UrbanLaneGraph~\cite{buchner_learning_2023}. The evaluation tiles from Pittsburgh are not considered for the values reported in the main paper due to their ground truth quality.
(P/R/$\mathrm{F}_\mathrm{1}$) denote precision, recall and $\mathrm{F}_\mathrm{1}$ score, IoU the rasterized Intersection over Union. BGFormer~\cite{blayney_bezier_2024} is abbreviated as BGF, LaneGNN~\cite{buchner_learning_2023} as LG and MapDreamer (Ours) as MD.
Note that LaneGNN is initialized using 178 ground truth poses for each global tile.}
\label{tab:quantitative_global_per_tile}
\resizebox{\textwidth}{!}{%
\begin{tabular}{@{}clcclclcclcclcclcclcc@{}}
\toprule
\multicolumn{1}{l}{} & City & \multicolumn{2}{c}{Austin} &  & Detroit &  & \multicolumn{2}{c}{Miami} &  & \multicolumn{2}{c}{Palo Alto} &  & \multicolumn{2}{c}{{\color[HTML]{CB0000} Pittsburgh}} &  & \multicolumn{2}{c}{Washington} &  &  &  \\ \cmidrule(lr){3-4} \cmidrule(lr){6-6} \cmidrule(lr){8-9} \cmidrule(lr){11-12} \cmidrule(lr){14-15} \cmidrule(lr){17-18}
\multicolumn{1}{r}{} & \multicolumn{1}{r}{Tile ID} & 40 & 83 &  & 165 &  & 185 & 194 &  & 43 & 62 &  & {\color[HTML]{CB0000} 5} & {\color[HTML]{CB0000} 36} &  & 46 & 55 &  & \multirow{-2}{*}{Mean} & \multirow{-2}{*}{\begin{tabular}[c]{@{}c@{}}Mean\\ (Reported)\end{tabular}} \\ \midrule
 & BGF & 0.097 & 0.277 &  & 0.269 &  & 0.404 & 0.469 &  & 0.311 & 0.416 &  & {\color[HTML]{CB0000} 0.210} & {\color[HTML]{CB0000} 0.299} &  & 0.526 & 0.402 &  & 0.335 & 0.352 \\
 & LG & \textbf{0.588} & 0.543 &  & 0.428 &  & \textbf{0.563} & 0.617 &  & 0.476 & \textbf{0.638} &  & {\color[HTML]{CB0000} 0.395} & {\color[HTML]{CB0000} \textbf{0.560}} &  & 0.506 & 0.546 &  & 0.533 & 0.545 \\
\multirow{-3}{*}{\begin{tabular}[c]{@{}c@{}}$\mathrm{GEO}_{1}$\\ $\mathrm{F}_1$\end{tabular}} & MD & 0.526 & \textbf{0.721} &  & \textbf{0.537} &  & 0.539 & \textbf{0.672} &  & \textbf{0.527} & 0.571 &  & {\color[HTML]{CB0000} \textbf{0.574}} & {\color[HTML]{CB0000} 0.411} &  & \textbf{0.592} & \textbf{0.572} &  & \textbf{0.567} & \textbf{0.584} \\ \midrule
 & BGF & 0.043 & 0.121 &  & 0.133 &  & 0.244 & 0.282 &  & 0.159 & 0.196 &  & {\color[HTML]{CB0000} 0.076} & {\color[HTML]{CB0000} 0.149} &  & 0.309 & 0.214 &  & 0.175 & 0.189 \\
 & LG & 0.349 & 0.309 &  & 0.113 &  & 0.340 & 0.375 &  & 0.241 & \textbf{0.349} &  & {\color[HTML]{CB0000} 0.174} & {\color[HTML]{CB0000} \textbf{0.235}} &  & 0.254 & 0.301 &  & 0.276 & 0.292 \\
\multirow{-3}{*}{\begin{tabular}[c]{@{}c@{}}$\mathrm{TOPO}_{1}$\\ $\mathrm{F}_1$\end{tabular}} & MD & \textbf{0.354} & \textbf{0.535} &  & \textbf{0.295} &  & \textbf{0.457} & \textbf{0.511} &  & \textbf{0.328} & 0.311 &  & {\color[HTML]{CB0000} \textbf{0.246}} & {\color[HTML]{CB0000} 0.173} &  & \textbf{0.433} & \textbf{0.388} &  & \textbf{0.366} & \textbf{0.401} \\ \midrule
 & BGF & 0.076 & 0.208 &  & 0.263 &  & 0.347 & 0.350 &  & 0.277 & 0.303 &  & {\color[HTML]{CB0000} 0.163} & {\color[HTML]{CB0000} 0.244} &  & 0.400 & 0.301 &  & 0.267 & 0.281 \\
 & LG & 0.392 & 0.374 &  & 0.313 &  & 0.405 & 0.420 &  & 0.336 & \textbf{0.458} &  & {\color[HTML]{CB0000} 0.333} & {\color[HTML]{CB0000} \textbf{0.423}} &  & 0.351 & 0.412 &  & 0.383 & 0.385 \\
\multirow{-3}{*}{$\mathrm{IoU}_1$} & MD & \textbf{0.394} & \textbf{0.555} &  & \textbf{0.455} &  & \textbf{0.429} & \textbf{0.512} &  & \textbf{0.415} & 0.422 &  & {\color[HTML]{CB0000} \textbf{0.463}} & {\color[HTML]{CB0000} 0.366} &  & \textbf{0.463} & \textbf{0.444} &  & \textbf{0.447} & \textbf{0.454} \\ \bottomrule
\end{tabular}%
}
\end{table}

\subsubsection{Qualitative Results.}
\begin{figure}[hp]
    \centering
    \begin{subfigure}[b]{0.49\linewidth}
        \centering
        \def\svgwidth{\linewidth} 
        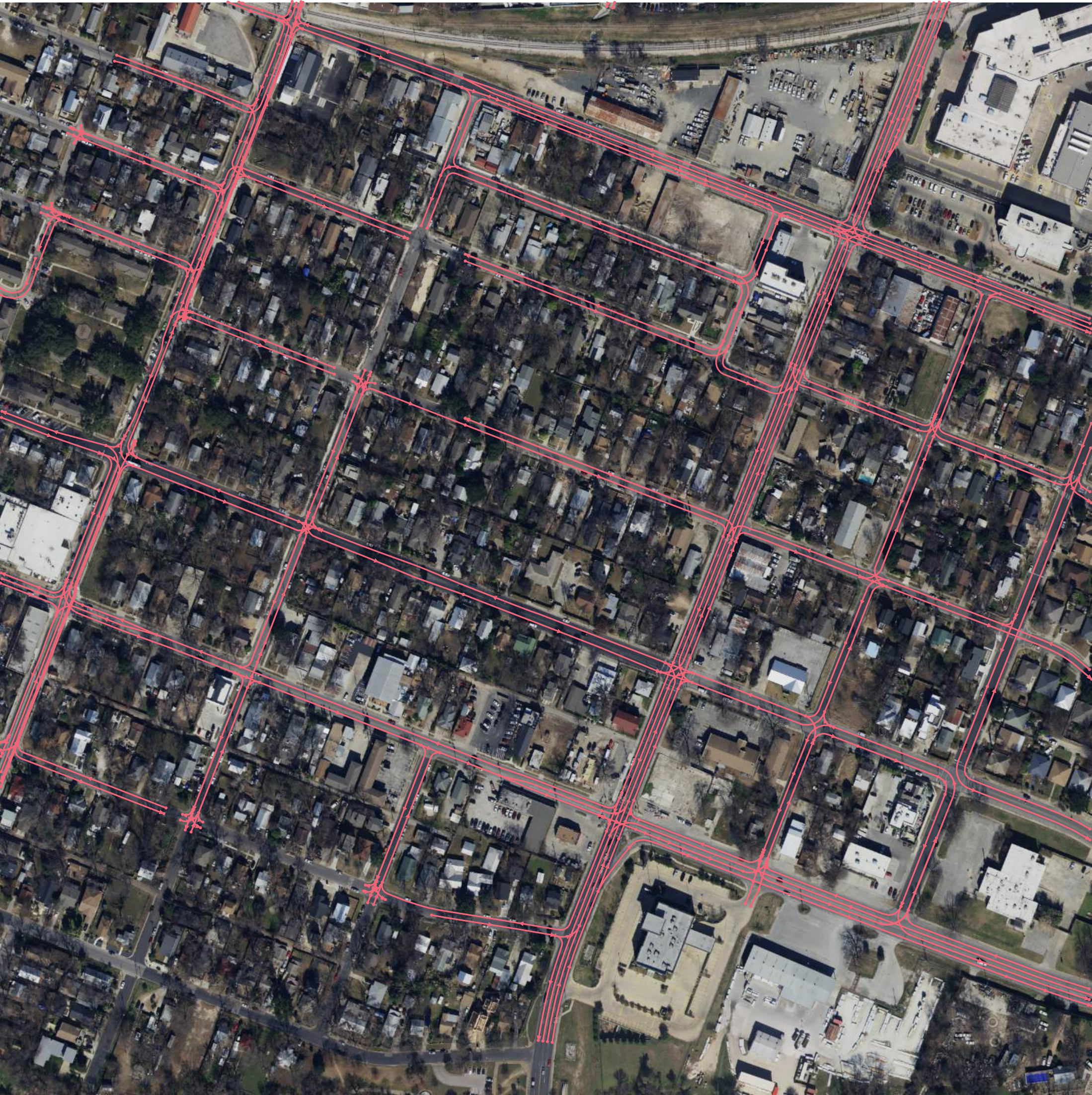
        \caption{Ground Truth}
        \label{fig:austin83_GT}
    \end{subfigure}
    \begin{subfigure}[b]{0.49\linewidth}
        \centering
        \def\svgwidth{\linewidth} 
        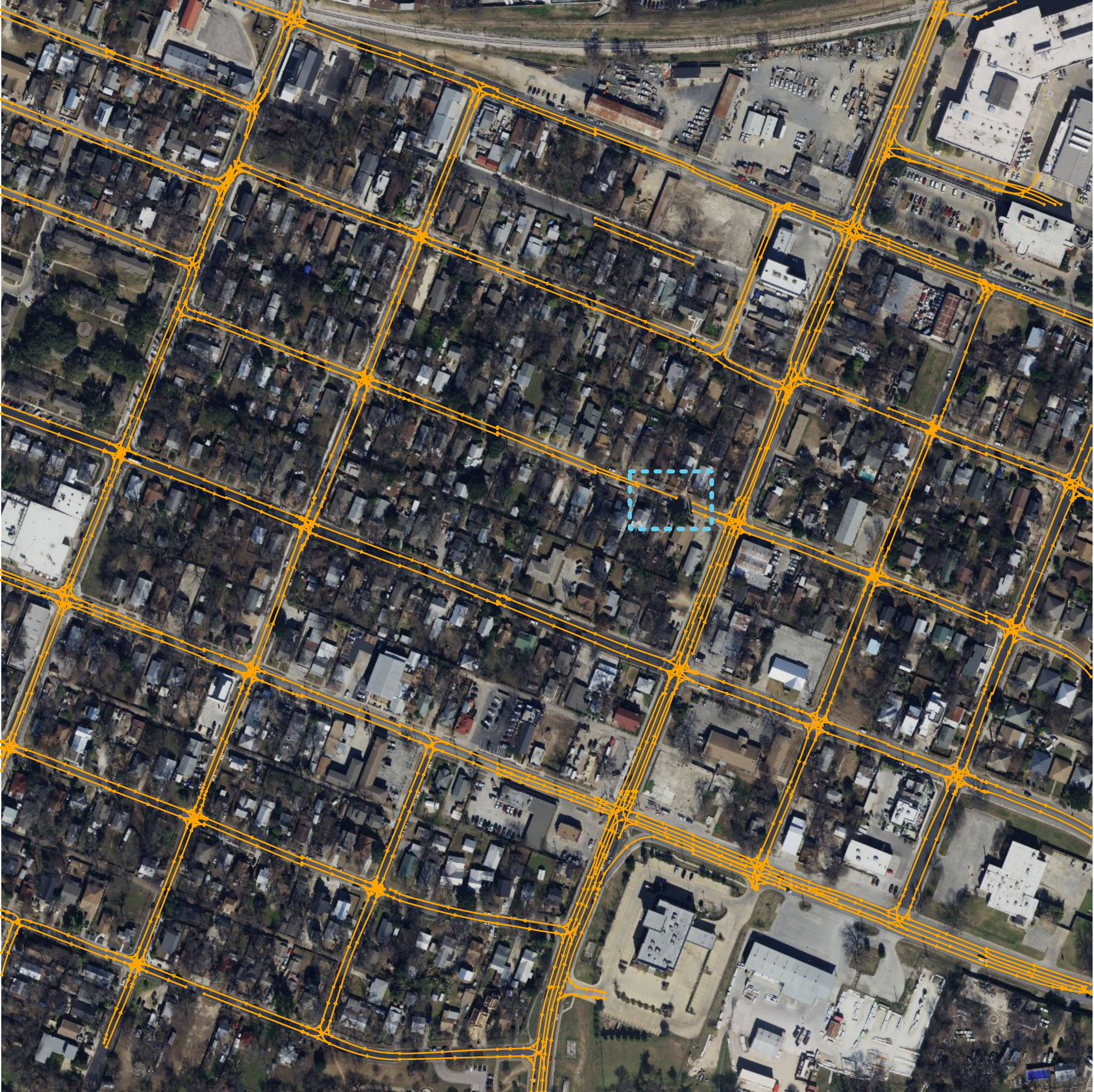
        \caption{MapDreamer (Ours)}
        \label{fig:austin83_MD}
    \end{subfigure}

    \medskip
    
    \begin{subfigure}[b]{0.49\linewidth}
        \centering
        \def\svgwidth{\linewidth} 
        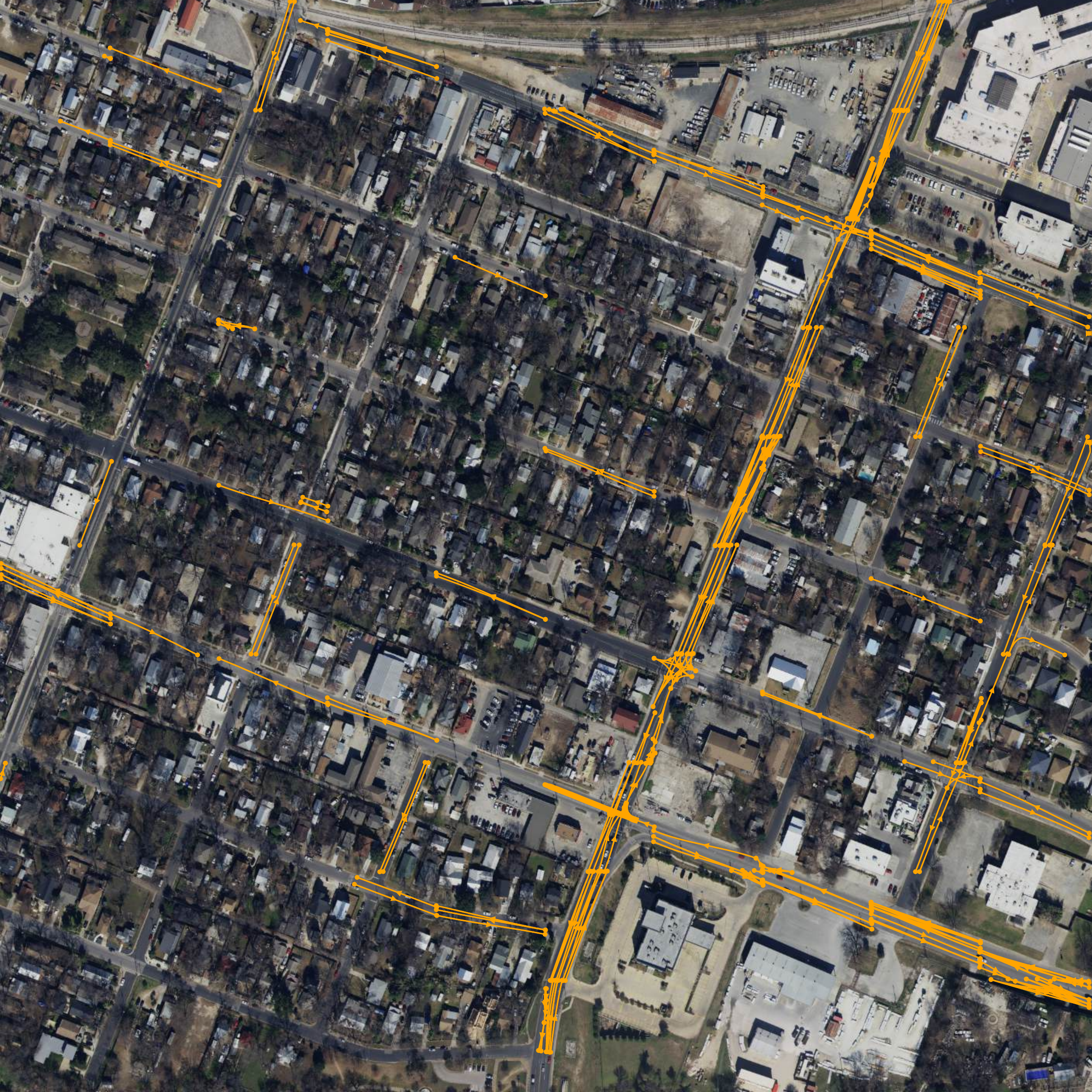
        \caption{BGFormer~\cite{blayney_bezier_2024}}
        \label{fig:austin83_BGF}
    \end{subfigure}
    \begin{subfigure}[b]{0.49\linewidth}
        \centering
        \def\svgwidth{\linewidth} 
        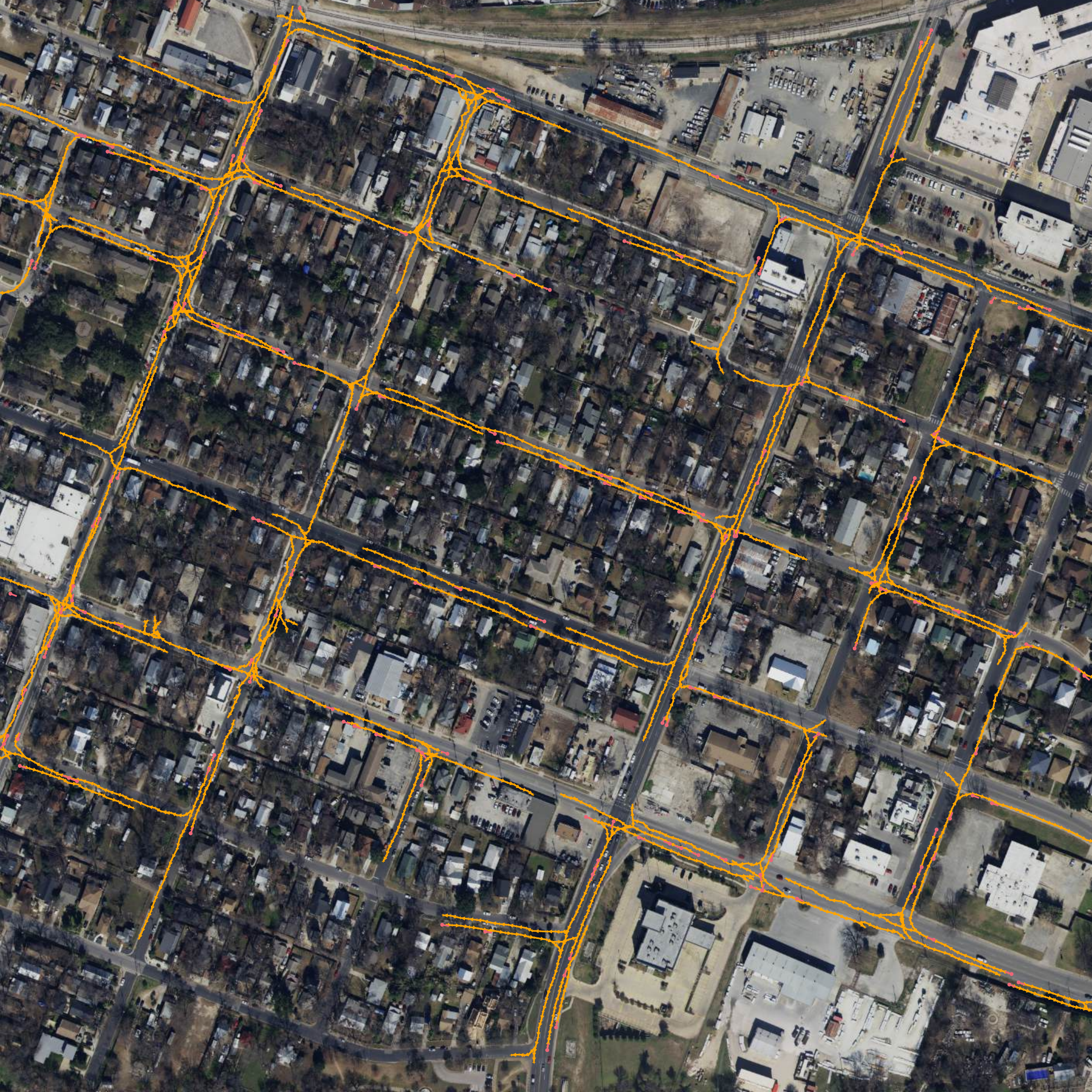
        \caption{LaneGNN~\cite{buchner_learning_2023}}
        \label{fig:austin83_LNN}
    \end{subfigure}

    \caption{Qualitative results for global graph generation on a full evaluation tile from Austin, Texas (UrbanLaneGraph~\cite{buchner_learning_2023} ID 83). A crop of this evaluation tile is also shown in the main paper. Subfigure \subref{fig:austin83_GT} shows the ground truth data in pink. Orange lines in \subref{fig:austin83_MD}, \subref{fig:austin83_BGF}, and \subref{fig:austin83_LNN} correspond to predicted lanes from the respective method. For LaneGNN (\subref{fig:austin83_LNN}), pink arrows indicate graph traversal initialization poses from ground truth. Note that LaneGNN and BGFormer discard predicted lanes with no ground truth information within a $50\,\mathrm{px}$ radius. Best viewed when zoomed in, the original resolution of $5000\,\mathrm{px}\times5000\,\mathrm{px}$ is reduced for visualization.}
    \label{fig:supplementary_global_austin_83}
\end{figure}

\begin{figure}[hp]
    \centering
    \begin{subfigure}[b]{0.49\linewidth}
        \centering
        \def\svgwidth{\linewidth} 
        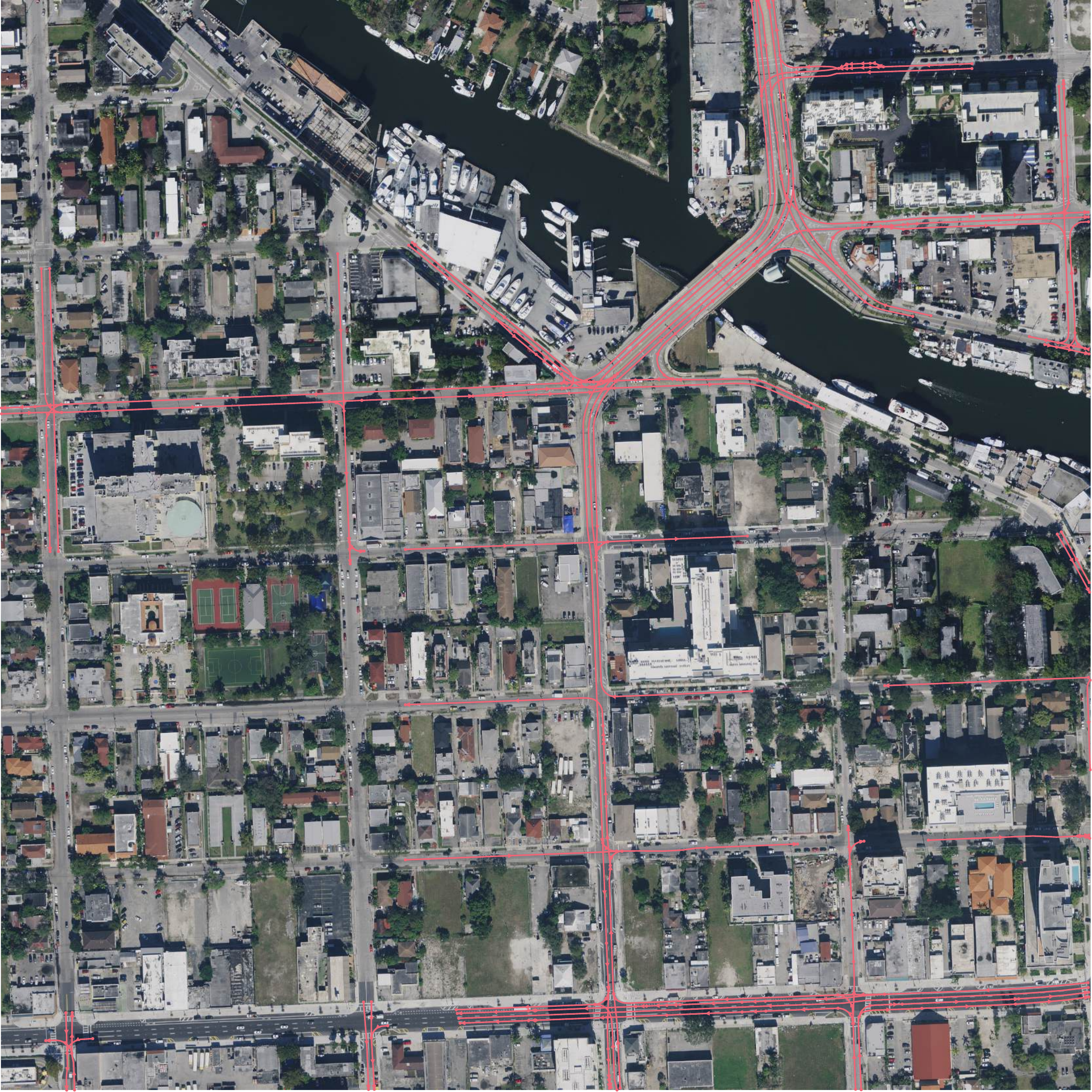
        \caption{Ground Truth}
        \label{fig:miami185_GT}
    \end{subfigure}
    \begin{subfigure}[b]{0.49\linewidth}
        \centering
        \def\svgwidth{\linewidth} 
        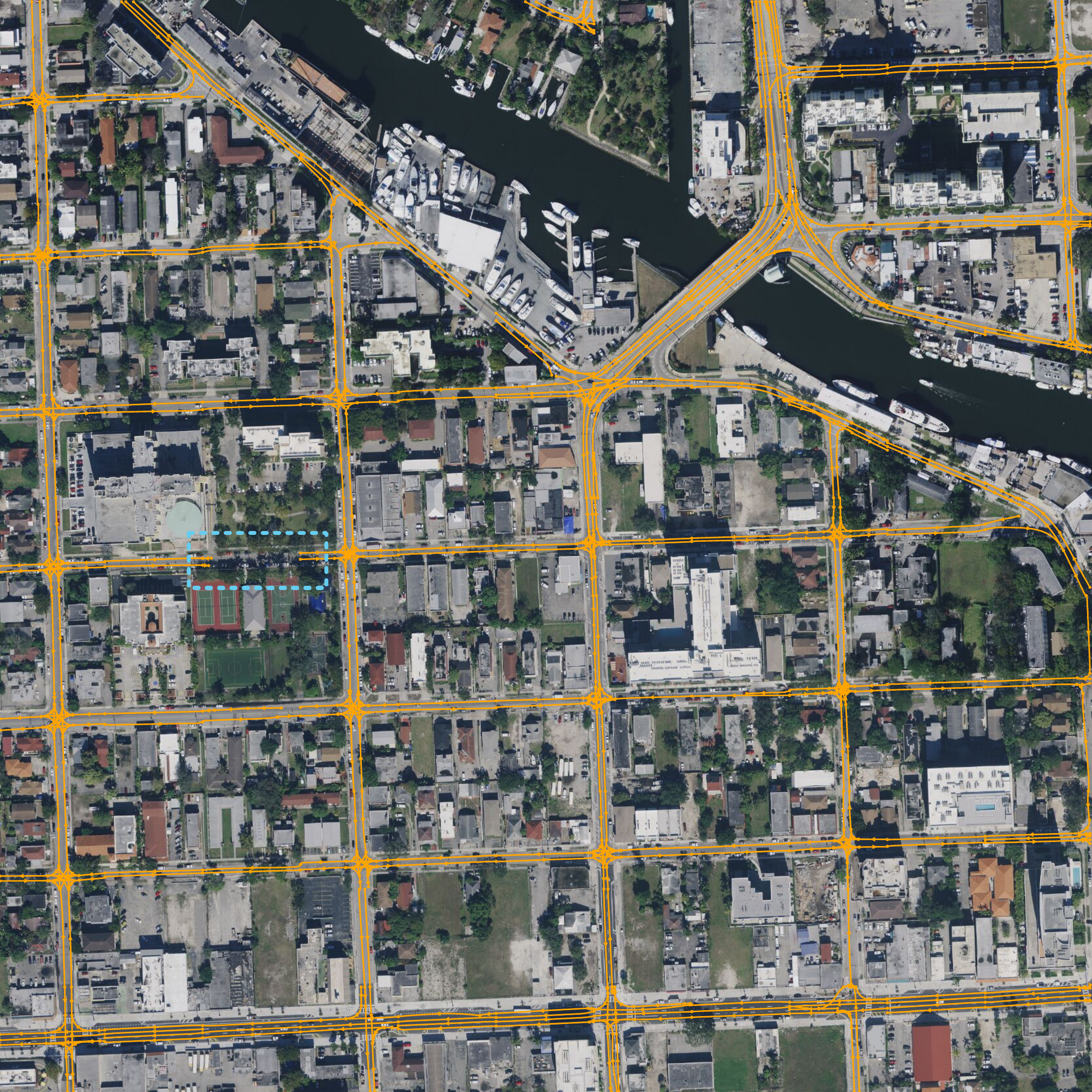
        \caption{MapDreamer (Ours)}
        \label{fig:miami185_MD}
    \end{subfigure}

    \medskip
    
    \begin{subfigure}[b]{0.49\linewidth}
        \centering
        \def\svgwidth{\linewidth} 
        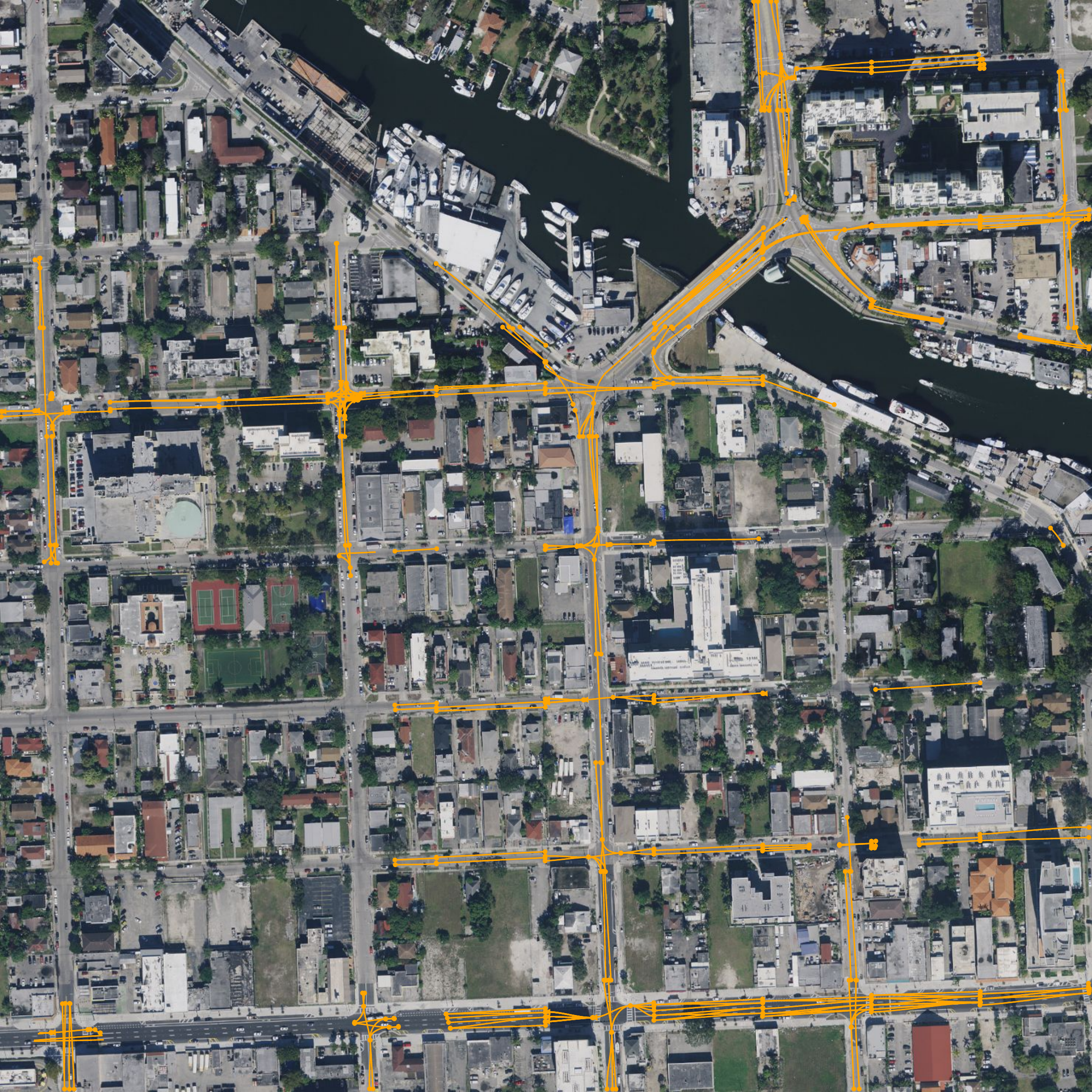
        \caption{BGFormer~\cite{blayney_bezier_2024}}
        \label{fig:miami185_BGF}
    \end{subfigure}
    \begin{subfigure}[b]{0.49\linewidth}
        \centering
        \def\svgwidth{\linewidth} 
        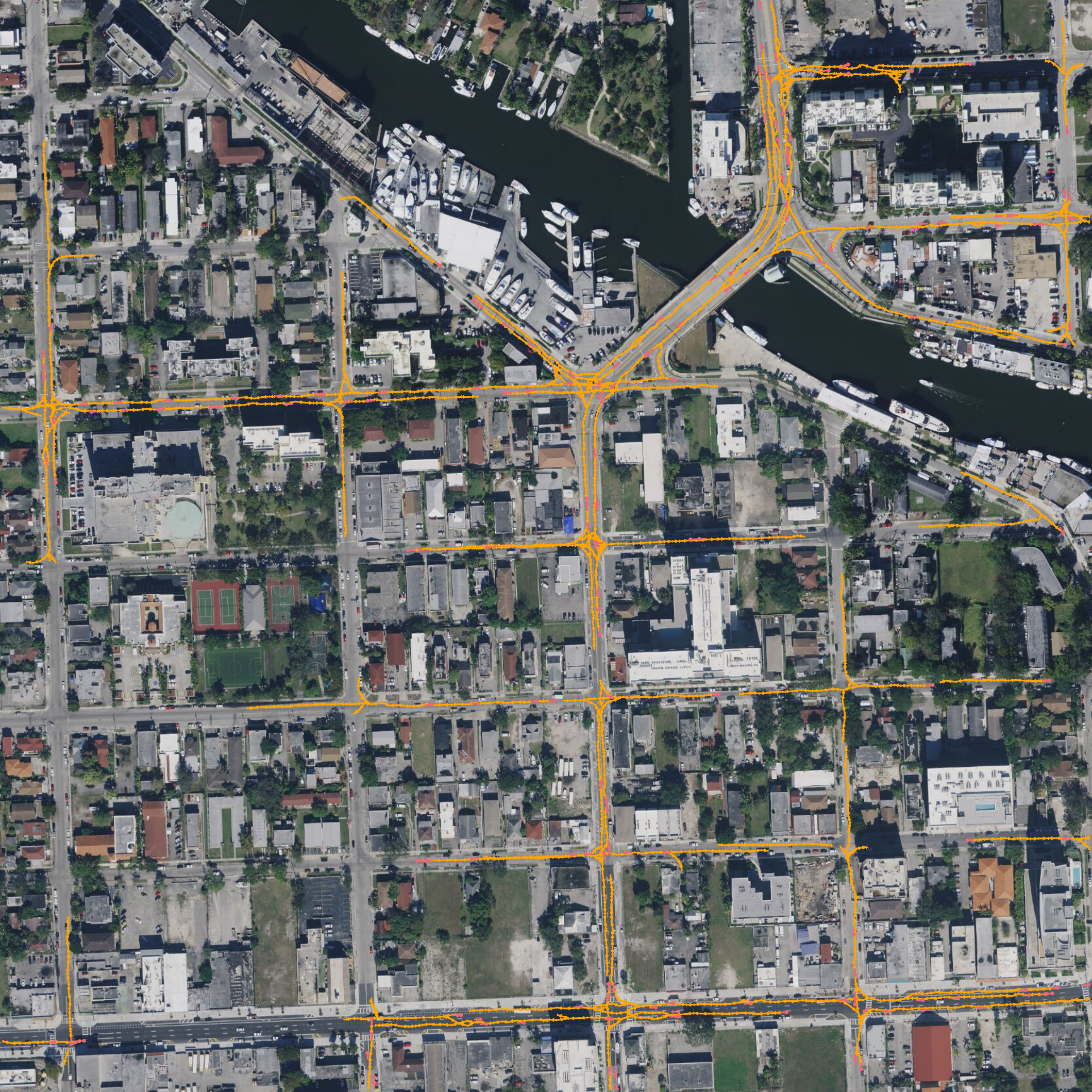
        \caption{LaneGNN~\cite{buchner_learning_2023}}
        \label{fig:miami185_LNN}
    \end{subfigure}

    \caption{Qualitative results for global graph generation on full evaluation tile from Miami (UrbanLaneGraph~\cite{buchner_learning_2023} ID 185). Subfigure \subref{fig:miami185_GT} displays the ground truth data in pink. Orange lines in \subref{fig:miami185_MD}, \subref{fig:miami185_BGF}, and \subref{fig:miami185_LNN} correspond to predicted lanes from the respective methods. For LaneGNN (\subref{fig:miami185_LNN}), pink arrows indicate graph traversal initialization poses sampled from ground truth. Note that LaneGNN and BGFormer discard predicted lanes with no ground truth information within a $50\,\mathrm{px}$ radius. Best viewed zoomed in, the original resolution of $5000\,\mathrm{px}\times5000\,\mathrm{px}$ is reduced for visualization.}
    \label{fig:supplementary_global_miami_185}
\end{figure}

\begin{figure}[hp]
    \centering
    \begin{subfigure}[b]{0.49\linewidth}
        \centering
        \def\svgwidth{\linewidth} 
        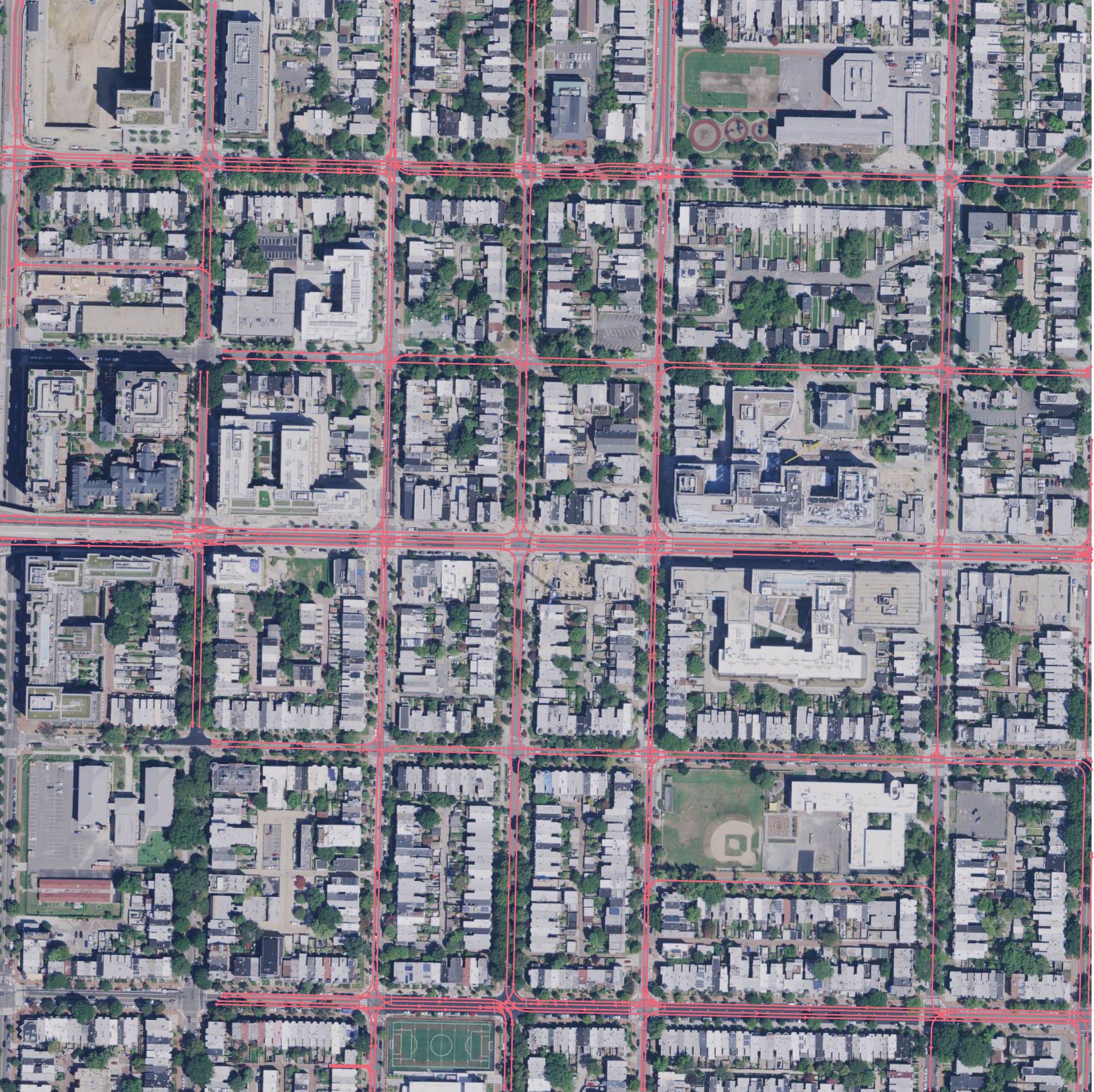
        \caption{Ground Truth}
        \label{fig:washington46_GT}
    \end{subfigure}
    \begin{subfigure}[b]{0.49\linewidth}
        \centering
        \def\svgwidth{\linewidth} 
        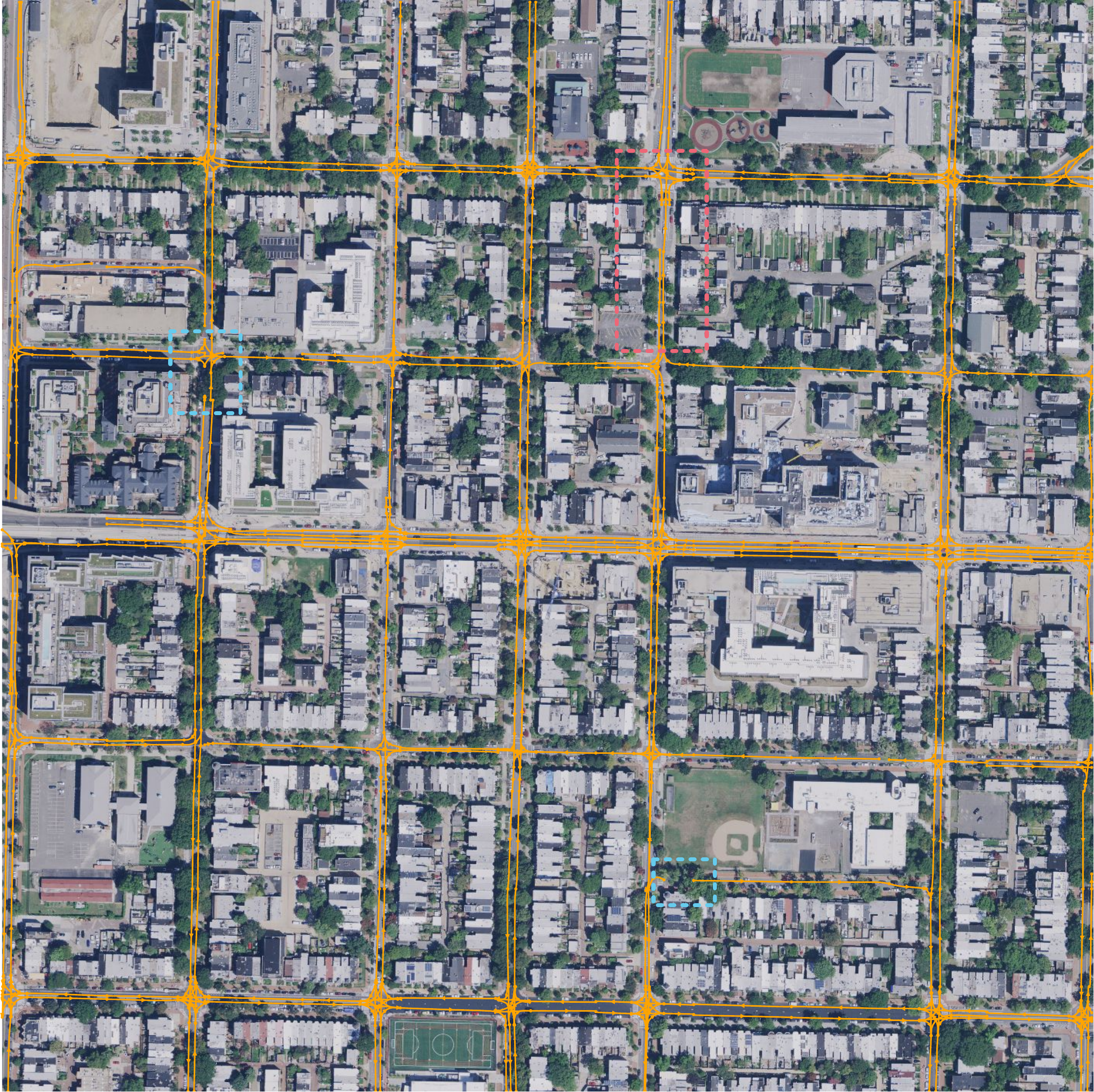
        \caption{MapDreamer (Ours)}
        \label{fig:washington46_MD}
    \end{subfigure}
    \medskip
    \begin{subfigure}[b]{0.49\linewidth}
        \centering
        \def\svgwidth{\linewidth} 
        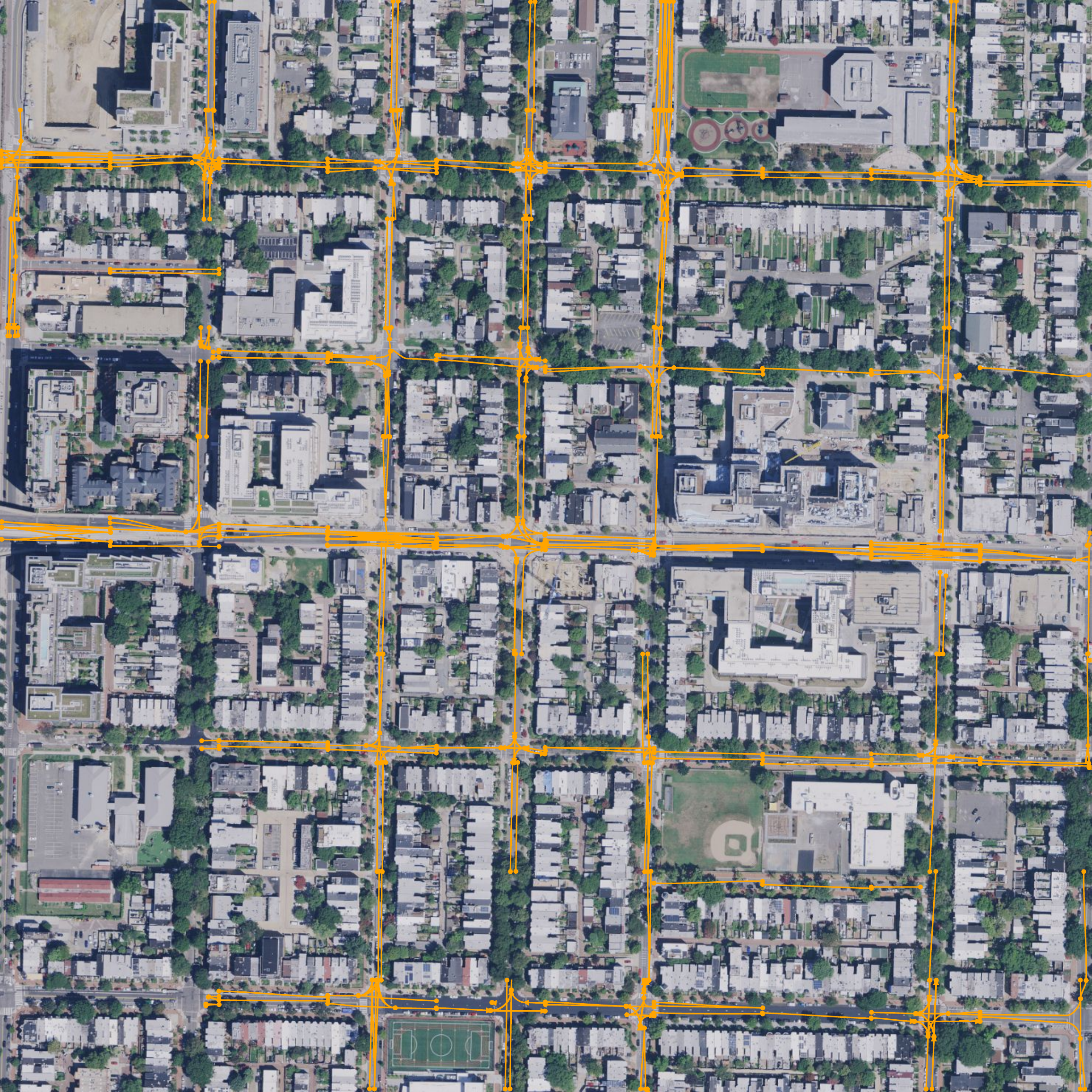
        \caption{BGFormer~\cite{blayney_bezier_2024}}
        \label{fig:washington46_BGF}
    \end{subfigure}
    \begin{subfigure}[b]{0.49\linewidth}
        \centering
        \def\svgwidth{\linewidth} 
        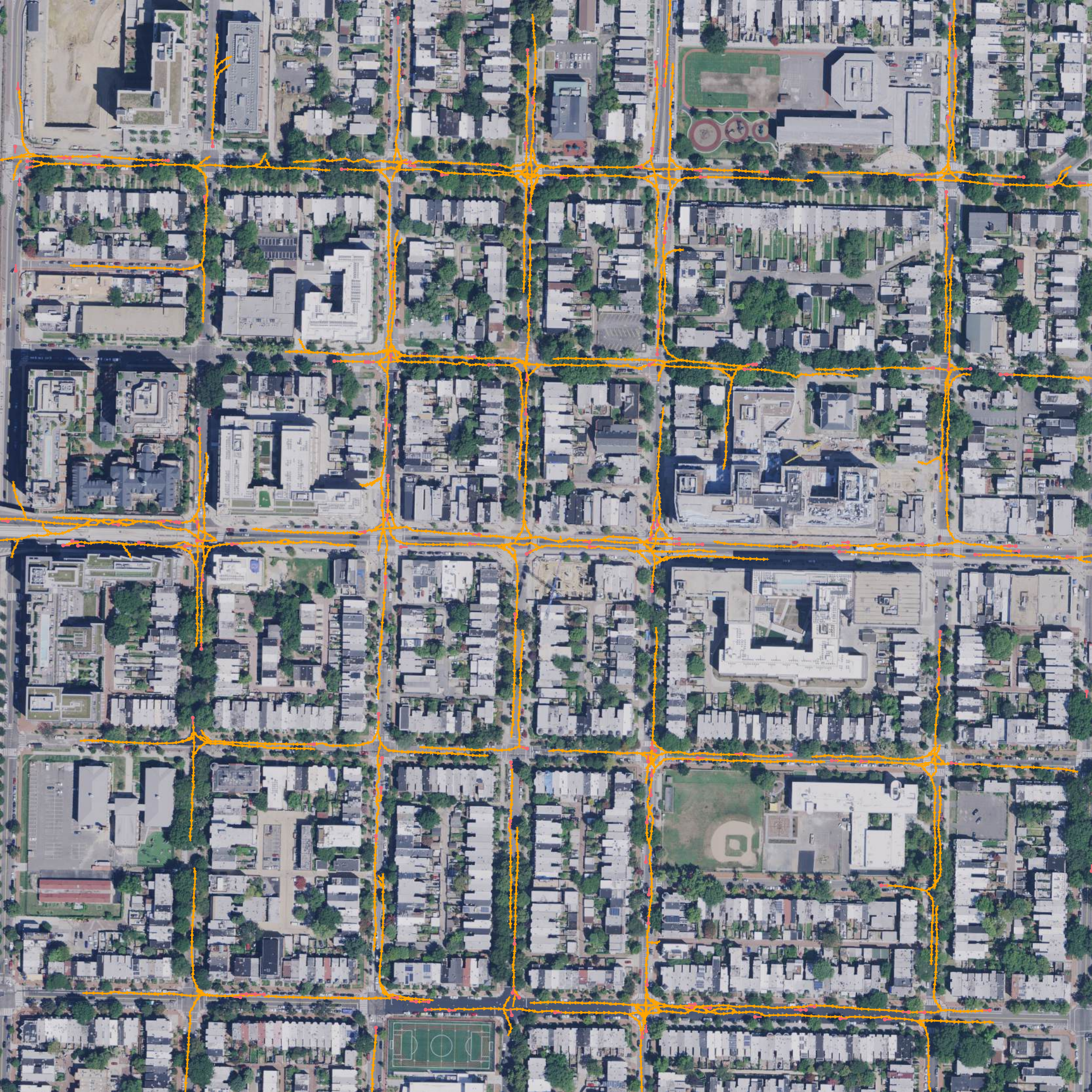
        \caption{LaneGNN~\cite{buchner_learning_2023}}
        \label{fig:washington46_LNN}
    \end{subfigure}

    \caption{Qualitative results for global graph generation on full evaluation tile from Washington (UrbanLaneGraph~\cite{buchner_learning_2023} ID 46). 
    Subfigure \subref{fig:washington46_GT} displays the ground truth data in pink. Orange lines in \subref{fig:washington46_MD}, \subref{fig:washington46_BGF}, and \subref{fig:washington46_LNN} correspond to predicted lanes from the respective method. For LaneGNN (\subref{fig:washington46_LNN}), pink arrows indicate graph traversal initialization poses from ground truth. Note that LaneGNN and BGFormer discard predicted lanes with no ground truth information within a $50\,\mathrm{px}$ radius. Best viewed zoomed in, the original resolution of $5000\,\mathrm{px}\times5000\,\mathrm{px}$ is reduced for visualization.}
    \label{fig:supplementary_global_washington_46}
\end{figure}

\Cref{fig:supplementary_global_austin_83}, \ref{fig:supplementary_global_miami_185} and \ref{fig:supplementary_global_washington_46} show qualitative results for MapDreamer on full global lane-graph evaluation tiles from UrbanLaneGraph~\cite{buchner_learning_2023} across three cities, namely Austin, Miami, and Washington.

We compare our results with BGFormer~\cite{blayney_bezier_2024} and LaneGNN~\cite{buchner_learning_2023}. For BGFormer and LaneGNN, lane predictions without ground-truth support within a radius of $50\,\mathrm{px}$ are discarded. For MapDreamer, we retain all predicted lanes for visualization, but apply the same correspondence threshold when computing quantitative metrics.

Qualitatively, MapDreamer performs best across all three evaluation tiles, with higher local geometric accuracy and more consistent large-scale connectivity. The local-to-global graph aggregation, implemented as a sliding-window strategy, preserves geometric and topological consistency across tile boundaries through boundary cross-attention.

\subsubsection{Failure Modes.}
Disconnected lane segments occur primarily in regions with strong occlusions caused by shadows, trees, or buildings. We highlight several such cases in
\cref{fig:supplementary_global_austin_83}\subref{fig:austin83_MD}, \cref{fig:supplementary_global_miami_185}\subref{fig:miami185_MD}, and \cref{fig:supplementary_global_washington_46}\subref{fig:washington46_MD} using \textcolor[HTML]{66CCEE}{cyan} dotted bounding boxes.

\begin{figure}
    \centering
    \begin{subfigure}[b]{0.49\linewidth}
        \centering
        \def\svgwidth{0.6\linewidth}
\begingroup%
  \makeatletter%
  \providecommand\color[2][]{%
    \errmessage{(Inkscape) Color is used for the text in Inkscape, but the package 'color.sty' is not loaded}%
    \renewcommand\color[2][]{}%
  }%
  \providecommand\transparent[1]{%
    \errmessage{(Inkscape) Transparency is used (non-zero) for the text in Inkscape, but the package 'transparent.sty' is not loaded}%
    \renewcommand\transparent[1]{}%
  }%
  \providecommand\rotatebox[2]{#2}%
  \newcommand*\fsize{\dimexpr\f@size pt\relax}%
  \newcommand*\lineheight[1]{\fontsize{\fsize}{#1\fsize}\selectfont}%
  \ifx\svgwidth\undefined%
    \setlength{\unitlength}{102.99964845bp}%
    \ifx\svgscale\undefined%
      \relax%
    \else%
      \setlength{\unitlength}{\unitlength * \real{\svgscale}}%
    \fi%
  \else%
    \setlength{\unitlength}{\svgwidth}%
  \fi%
  \global\let\svgwidth\undefined%
  \global\let\svgscale\undefined%
  \makeatother%
  \begin{picture}(1,1.86408393)%
    \lineheight{1}%
    \setlength\tabcolsep{0pt}%
    \put(0,0){\includegraphics[width=\unitlength,page=1]{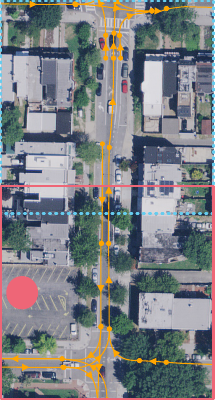}}%
    \put(0.07874922,0.46718459){\color[rgb]{0.65882353,0.03921569,0.11764706}\makebox(0,0)[lt]{\lineheight{1.20000005}\smash{\begin{tabular}[t]{l}1\end{tabular}}}}%
    \put(0,0){\includegraphics[width=\unitlength,page=2]{washington46_top_to_bottom.pdf}}%
    \put(0.07874922,1.33126832){\color[rgb]{0.01176471,0.34509804,0.45490196}\makebox(0,0)[lt]{\lineheight{1.20000005}\smash{\begin{tabular}[t]{l}2\end{tabular}}}}%
  \end{picture}%
\endgroup%

        \caption{Bottom-to-top inference (default)}
        \label{fig:bottom_to_top}
    \end{subfigure}
    \begin{subfigure}[b]{0.49\linewidth}
        \centering
        \def\svgwidth{0.6\linewidth}
\begingroup%
  \makeatletter%
  \providecommand\color[2][]{%
    \errmessage{(Inkscape) Color is used for the text in Inkscape, but the package 'color.sty' is not loaded}%
    \renewcommand\color[2][]{}%
  }%
  \providecommand\transparent[1]{%
    \errmessage{(Inkscape) Transparency is used (non-zero) for the text in Inkscape, but the package 'transparent.sty' is not loaded}%
    \renewcommand\transparent[1]{}%
  }%
  \providecommand\rotatebox[2]{#2}%
  \newcommand*\fsize{\dimexpr\f@size pt\relax}%
  \newcommand*\lineheight[1]{\fontsize{\fsize}{#1\fsize}\selectfont}%
  \ifx\svgwidth\undefined%
    \setlength{\unitlength}{103.00000529bp}%
    \ifx\svgscale\undefined%
      \relax%
    \else%
      \setlength{\unitlength}{\unitlength * \real{\svgscale}}%
    \fi%
  \else%
    \setlength{\unitlength}{\svgwidth}%
  \fi%
  \global\let\svgwidth\undefined%
  \global\let\svgscale\undefined%
  \makeatother%
  \begin{picture}(1,1.8640842)%
    \lineheight{1}%
    \setlength\tabcolsep{0pt}%
    \put(0,0){\includegraphics[width=\unitlength,page=1]{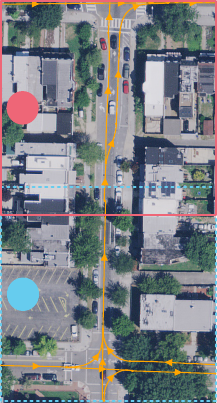}}%
    \put(0.07989651,1.33126871){\color[rgb]{0.65882353,0.03921569,0.11764706}\makebox(0,0)[lt]{\lineheight{1.20000005}\smash{\begin{tabular}[t]{l}1\end{tabular}}}}%
    \put(0.08166116,0.46718501){\color[rgb]{0.01176471,0.34509804,0.45490196}\makebox(0,0)[lt]{\lineheight{1.20000005}\smash{\begin{tabular}[t]{l}2\end{tabular}}}}%
  \end{picture}%
\endgroup%

        \caption{Top-to-bottom inference}
        \label{fig:top_to_bottom}
    \end{subfigure}
    \caption{Visualization of bottom-to-top and top-to-bottom inference strategies for a crop in the Washington 46 evaluation tile corresponding to the pink bounding box in \cref{fig:supplementary_global_washington_46}\subref{fig:washington46_MD}.}
    \label{fig:global_inference_strategy_failure_case}
\end{figure}

In some cases, incorrect outputs arise from the selected tiling order, as shown in the \textcolor[HTML]{EE6677}{pink dotted bounding box} in \cref{fig:supplementary_global_washington_46}\subref{fig:washington46_MD} and magnified in \cref{fig:global_inference_strategy_failure_case}.
Under the default inference order from bottom-left to top-right, the lower tile predicts a northbound two-way road segment (see \textcolor[HTML]{EE6677}{pink frame} in \cref{fig:global_inference_strategy_failure_case}\subref{fig:bottom_to_top}). 
However, the visual evidence indicating a one-way road, specifically the arrow markings on the pavement, is only visible in the upper tile.

Because the lane prediction in the second tile (\textcolor[HTML]{66CCEE}{cyan frame}) is conditioned on the previously generated bottom tile, it inherits the incorrect direction. By contrast, under the top-to-bottom inference strategy \cref{fig:global_inference_strategy_failure_case}\subref{fig:top_to_bottom}, the one-way road is correctly identified in the upper tile (\textcolor[HTML]{EE6677}{pink frame}). The bottom tile (\textcolor[HTML]{66CCEE}{cyan frame}), conditioned on the single lane facing north, then continues the one-way road until the next intersection.
This suggests that globally adaptive inference schedules may further improve large-scale consistency and represent an interesting direction for future work.

\subsection{Ablations}
\subsubsection{Lane Cardinality and Ghost Lane Latents.}
\begin{figure}[b]
    \centering
    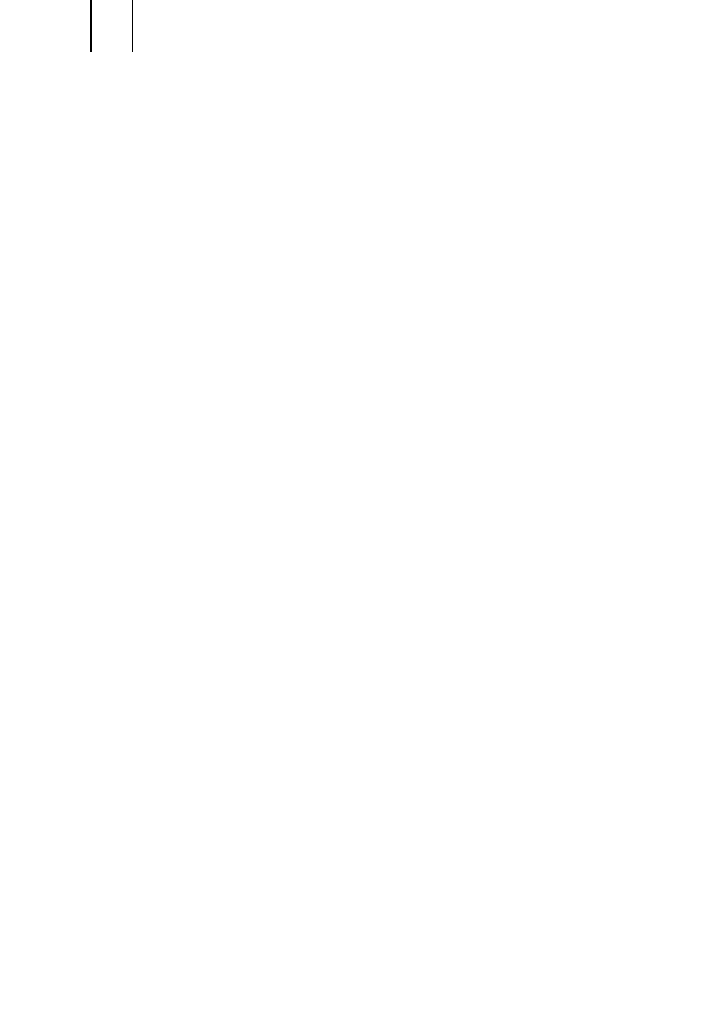
    \caption{Qualitative results for the ablation on the use of ghost lane latents (GL) and the lane cardinality module (LC) on the local lane graph generation task. The latent space is used for all models through the variational autoencoder (VAE). We highlight missing lanes in the no ghost lane latent ablation using \textcolor[HTML]{EE6677}{pink} ellipses. Gray boxes show the predicted number of lanes through the lane cardinality module (LC) and the final number of lanes predicted using the lane existence head (LE).}
    \label{fig:abl_gl_cardinality}
\end{figure}
\Cref{fig:abl_gl_cardinality} shows qualitative results for four local tiles from our evaluation split across different cities.
We compare our full MapDreamer model (first column) with variants presented in the ablation studies of the main paper. The second column shows the results of MapDreamer without ghost lane latents. 
The full model handles both overestimation (first row) and underestimation (second to fourth row) of the lane count by the lane cardinality module more robustly than the ablated variants.

When the lane cardinality module overestimates the number of lanes, the diffusion model can map surplus queries toward ghost-lane latents during denoising, which helps the lane existence head to suppress invalid lanes.
When ghost lane latents are removed, the lane existence head seems to strictly rely on the estimation from the lane cardinality module (note that, for every tile, the estimate from lane existence head and lane cardinality module match).

In cases where the lane cardinality module underestimates the lane count, the model architecture stripped of the ghost lane latents has no way to correct this estimate. For the second and third rows, the full MapDreamer diffusion model is able to fix the underestimation through the use of one spare ghost lane latent, while the model without ghost latents misses a turning lane, highlighted using a pink ellipse.
For the complex intersection in the fourth row, the difference becomes even more apparent, as three additional ghost-lane latent queries are used in the diffusion process, improving the predicted geometry and topology layout significantly.

The results shown in the third column indicate the importance of the lane cardinality module in our inference pipeline. When removed, the predictions become much more spurious, overestimating the number of lanes for every displayed evaluation sample. In addition to higher inference cost from the diffusion model always denoising $N_{\max}$ lanes, the generated maps are of substantially reduced quality, matching the quantitative results presented in the main paper.

\subsubsection{Vectorized Diffusion Model.}
\begin{figure}[b]
    \centering
    \def\svgwidth{\linewidth} 
    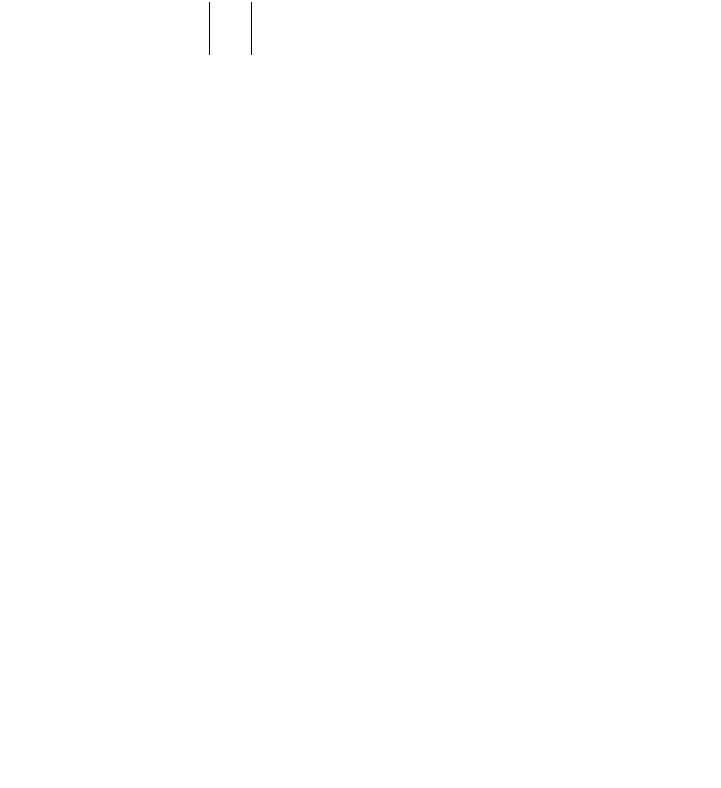
    \caption{Qualitative results for the ablation on latent diffusion using the variational autoencoder (VAE) compared to vectorized diffusion without the use of ghost lane latents (GL). We highlight degraded geometric endpoint alignment using \textcolor[HTML]{EE6677}{pink} ellipses. Gray boxes show the number of lanes predicted by the lane-cardinality module (LC) and the final number of lanes predicted using the lane existence head (LE).}
    \label{fig:abl_vae}
\end{figure}

In the main paper, we compare the latent diffusion formulation in MapDreamer with a baseline that performs diffusion directly in vectorized coordinate space, omitting the VAE encoder $E_\phi$ and decoder $\mathcal{D}_\gamma$ entirely.
Here, we extend this evaluation with qualitative results for three local graph generation samples from our validation data split in \cref{fig:abl_vae}. While direct vectorized diffusion handles simple scenarios reasonably well (see first row in \cref{fig:abl_vae}), its performance degrades significantly with increased scene complexity (see the second and third row in \cref{fig:abl_vae}).
In particular, geometric endpoint alignment is less accurate, as highlighted with pink ellipses in \cref{fig:abl_vae}. Further, the lane geometries appear less smooth and noisier.
These observations are consistent with the quantitative trends in the main paper and further support the latent diffusion design, where the autoencoder provides a learned structural prior over lane-level maps.

\section{Implementation Details}

We provide an overview of key model parameters used during training in \cref{tab:hyperparameters}. In addition, \cref{tab:training_schedule} summarizes the optimization settings for the autoencoder and latent diffusion model.

\begin{table}
\centering
\caption{Overview of model parameters used for the autoencoder and latent diffusion model architectures.}
\label{tab:hyperparameters}
\begin{tabular}{@{}lcc@{}}
\toprule
\multicolumn{3}{c}{Autoencoder} \\ \midrule
Number of encoder blocks $N_\mathrm{E}$ & \multicolumn{2}{c}{2} \\
Number of decoder blocks $N_\mathrm{D}$ & \multicolumn{2}{c}{2} \\
Hidden dimension $d_\mathrm{l}$ lane geometry & \multicolumn{2}{c}{512} \\
Hidden dimension $d_\mathrm{conn}$ lane connectivity & \multicolumn{2}{c}{64} \\
Latent dimension lane $K_\mathrm{l}$ & \multicolumn{2}{c}{24} \\
Reconstruction loss weight $\lambda_{\ell_1}$ & \multicolumn{2}{c}{10} \\
Reconstruction loss weight $\lambda_\mathrm{conn}$ & \multicolumn{2}{c}{10} \\
Lane endpoint loss weight $\lambda_\mathrm{end}$ & \multicolumn{2}{c}{10} \\ \midrule
\multicolumn{1}{c}{Latent Diffusion Model (LDM)} &  &  \\ \midrule
 & Default & Small \\
Number of decoder blocks & 2 & 2 \\
Hidden dimension $d_\mathrm{l}$ lane geometry & 2048 & 512 \\ \bottomrule
\end{tabular}
\end{table}

\begin{table}
\centering
\caption{Training schedule parameters used for autoencoder and latent diffusion model (LDM) training.}
\label{tab:training_schedule}
\begin{tabular}{@{}lcc@{}}
\toprule
\multicolumn{1}{c}{} & Autoencoder & LDM \\ \midrule
Optimizer & AdamW & AdamW \\
Learning rate & $2 \times 10^{-4}$ & $2 \times 10^{-4}$ \\
Weight decay & $1 \times 10^{-4}$ & $1 \times 10^{-4}$ \\
Total batch size & 96 & 288 \\
GPUs & 1 $\times$ V100 & 2 $\times$ V100 \\ \bottomrule
\end{tabular}
\end{table}

\end{document}